\documentclass[sigconf,authorversion]{acmart}

\setcopyright{none}
\renewcommand\footnotetextcopyrightpermission[1]{}
\acmDOI{}
\acmISBN{}
\acmConference[Preprint]{Preprint}{October 2026}{Shanghai, China}

\usepackage{booktabs}
\usepackage{nicefrac}
\usepackage{algorithm}
\usepackage{algorithmic}
\usepackage{amsmath}

\usepackage{amsthm}
\usepackage[capitalise,noabbrev]{cleveref}
\usepackage{graphicx}
\usepackage{subcaption}
\usepackage{multirow}
\usepackage{siunitx}
\usepackage{array}
\newcolumntype{L}[1]{>{\raggedright\arraybackslash}p{#1}}
\newcommand{\new}[1]{\textcolor{black}{#1}}

\newcommand{\fc}{\mathtt{w}}
\newcommand{\lc}{\mathfrak{l}}
\newcommand{\nc}{w}
\usepackage{bm}
\usepackage{enumitem}\usepackage{etoolbox}\definecolor{bestcellblue}{RGB}{225,240,255}\newcommand{\besthighlight}[1]{\begingroup\setlength{\fboxsep}{1.5pt}\colorbox{bestcellblue}{\strut\bfseries #1}\endgroup}\AtBeginEnvironment{tabular}{\let\textbf\besthighlight}\AtBeginEnvironment{tabular*}{\let\textbf\besthighlight}
\newcommand{\rev}[1]{\textcolor{black}{#1}}

\title{\rev{KITINet: KInetic Theory Inspired Inter-Channel Information Exchange for Residual Networks}}
\author{Mingquan Feng}
\authornote{These authors contributed equally to this work.}
\affiliation{%
  \institution{Shanghai Jiao Tong University}
  \city{Shanghai}
  \country{China}
}
\email{fengmingquan@sjtu.edu.cn}

\author{Yifan Fu}
\authornotemark[1]
\affiliation{%
  \institution{Shanghai Jiao Tong University}
  \city{Shanghai}
  \country{China}
}

\author{Tongcheng Zhang}
\authornotemark[1]
\affiliation{%
  \institution{Shanghai Jiao Tong University}
  \city{Shanghai}
  \country{China}
}

\author{Yu Jiang}
\affiliation{%
  \institution{Shanghai Jiao Tong University}
  \city{Shanghai}
  \country{China}
}

\author{Yixin Huang}
\affiliation{%
  \institution{Shanghai Jiao Tong University}
  \city{Shanghai}
  \country{China}
}

\author{Junchi Yan}
\authornote{Corresponding author.}
\affiliation{%
  \institution{Shanghai Jiao Tong University}
  \city{Shanghai}
  \country{China}
}
\email{yanjunchi@sjtu.edu.cn}
\renewcommand{\shortauthors}{Feng et al.}

\begin{document}

\theoremstyle{definition}
\newtheorem{assumption}[theorem]{Assumption}
\theoremstyle{remark}
\newtheorem{remark}[theorem]{Remark}

\begin{abstract}
\rev{Residual connections are foundational to modern neural networks, yet the standard additive update provides no explicit mechanism for structured inter-channel interaction. We introduce KITINet (KInetics Theory Inspired Network), a training-time module with no additional trainable parameters that augments residual updates with stochastic pairwise feature interactions inspired by collision sampling in kinetic theory. KITINet reshapes channel groups as particles and mixes their features according to relative distance and velocity, while leaving the underlying residual layer trainable in the usual way. The module is disabled at inference, so the deployed model is identical to the original backbone and incurs no additional inference cost. Across language pre-training and fine-tuning, image classification, and PDE operator learning, KITINet demonstrates broad and robust performance gains across diverse tasks and architectures. We further observe enhanced parameter condensation across several settings and develop a simplified stochastic analysis that provides a mechanistic account of how collision-inspired interactions can promote condensation dynamics.}
\end{abstract}

\maketitle

\section{Introduction}
\label{sec:intro}

% Residual connections enable training of deep architectures by alleviating vanishing gradients. From ResNets \citep{he2016deep} to Transformers \citep{vaswani2017attention}, skip connections underpin state-of-the-art models. Yet their design has remained heuristic: standard skip connections propagate features via simple additive operations, leaving substantial room for structured inter-channel information exchange.

\rev{Residual connections have become a central mechanism for training deep networks, providing short information and gradient paths that stabilize feature propagation across many layers. From ResNets~\citep{he2016deep} in vision to Transformers~\citep{vaswani2017attention} in language modeling, they underpin a wide range of modern architectures. Yet the standard update combines the identity stream and newly computed features through fixed element-wise addition; it neither distinguishes channel groups nor explicitly controls how information is exchanged among them. Dynamical-systems interpretations~\citep{chen2018neural,norcliffe2020second,norcliffe2021neural} recast layer-wise propagation as numerical evolution, while PDE-based, Hamiltonian, and other physics-inspired networks~\citep{long2018pde,long2019pde,toth2019hamiltonian,greydanus2019hamiltonian,schuetz2022combinatorial} show how mathematical structure can guide architecture design. These perspectives motivate richer residual dynamics, but they largely retain deterministic, collision-free propagation rather than stochastic pairwise interactions between feature groups.}

\rev{Recent work has begun to treat residual pathways as structured communication mechanisms rather than passive additive shortcuts. Dit-ResNet~\citep{liu2024dendritic} augments residual blocks with dendritic integration to model data-dependent feature correlations, while COIN~\citep{wang2025convection} introduces convection--diffusion transformations into residual architectures. mHC~\citep{xie2026mhc} expands the residual stream and constrains learned cross-stream mixing to preserve identity mapping and stabilize large-scale training. These methods enrich residual communication through nonlinear integration, physical transformations, or constrained routing, but leave stochastic pairwise mixing under kinetic invariants unexplored.}

\rev{We introduce KITINet, a training-time residual module that reshapes feature channels as particles and applies distance- and relative-velocity-dependent stochastic pairwise interactions inspired by BTE and DSMC. Its symmetric collision rule conserves pairwise kinetic energy, providing a physical invariant analogous in structural role to the identity-mapping constraint in mHC. Through this constrained collision process, KITINet redistributes information across feature groups and enriches the residual update. At inference, KITINet reduces exactly to standard residual addition. Across several architectures, the resulting training dynamics are associated with stronger parameter condensation in our measurements, which helps explain the empirical gains.}
% \new{Unlike mHC, which enforces stability via an explicit constrained mixing matrix, KITINet achieves adaptive inter-channel refinement implicitly through energy-preserving collision-driven dynamics without introducing additional learnable routing variables.} 
\rev{We evaluate KITINet on language pre-training and fine-tuning, image classification, and PDE operator learning. Our contributions are:}
\begin{itemize}
    \item \rev{We introduce a training-time residual regularizer with no additional trainable parameters, based on structured stochastic pairwise feature interactions inspired by DSMC collision sampling; it is disabled at inference and therefore incurs no additional inference cost.}
    \item \rev{We provide a physics-motivated interpretation of parameter condensation, along with a simplified theoretical model that provides intuition for condensation-like dynamics.}
    \item \rev{We demonstrate consistent performance gains over baselines on language tasks, image classification, and PDE operator learning with zero additional inference cost.}
\end{itemize}
% \rev{\textbf{Positioning.} KITINet is closest to training-time stochastic regularization. Unlike feature masking such as Dropout~\citep{JMLR:v15:srivastava14a}, it perturbs the residual update through structured pairwise mixing and preserves every feature group in the update. Its novelty lies in this interaction rule and its compatibility with an unchanged inference graph, not in claiming a faithful physical simulation.}

\section{Related Work}
\label{sec:related}
\textbf{Residual Learning and Dynamical Systems.}
ResNet \citep{he2016deep} introduced residual connections to mitigate vanishing gradients in deep networks. Subsequent studies reinterpreted residual networks through dynamical systems theory, with neural ODEs \citep{chen2018neural} modeling continuous-depth networks as ordinary differential equations (ODEs). ODE-RNN \citep{rubanova2019latent} simulates continuous dynamics of hidden states in RNNs. Neural-controlled differential equations \citep{kidger2020neural} extend this framework to incorporate control mechanisms, enabling adaptive feature propagation. Other extensions include second-order residuals \citep{norcliffe2020second} and flow models \citep{bilovs2021neural}. While these works provide valuable insights into the dynamics of residual learning, they primarily focus on deterministic, collision-free dynamics, neglecting the stochastic interactions and energy dissipation mechanisms inherent in real-world particle systems.

\textbf{Physics-Inspired Network Architectures}
Recent efforts have integrated physical principles into neural architectures to enhance interpretability and data efficiency. Hamiltonian networks \citep{greydanus2019hamiltonian} preserve energy conservation laws, and Lagrangian networks \citep{toth2019hamiltonian} derive updates from variational principles. \rev{PDE-inspired models, such as PDE-GCN~\citep{eliasof2021pde} and the PDE-Net family~\citep{long2018pde,long2019pde}, parameterize spatiotemporal evolution using PDE-structured operators.} Closest to our work, \citet{wang2025convection} proposed a convection-diffusion network (COIN), which incorporates diffusion layers after the ResNet architecture. But their formulation lacks explicit ties to residual learning or parameter condensation. Critically, while the above frameworks borrow mathematical structures from physics, they do not simulate collisional processes or exploit thermodynamic relaxation for network sparsity.

\section{Preliminaries}
\subsection{Kinetic Theory and the Boltzmann Transport Equation}

Kinetic theory describes macroscopic particle system behavior through microscopic collision dynamics. The central object is the density function $f(\boldsymbol{x}, \boldsymbol{p}, t)$ in $7$-dimensional phase space, defined via $dN = f(\boldsymbol{x}, \boldsymbol{p}, t) \, d^3 \boldsymbol{x} \, d^3 \boldsymbol{p}$. Assuming Hamiltonian dynamics under external force $F_{ex}$, $f$ satisfies the \textbf{Boltzmann Transport Equation (BTE)}~\citep{boltzmann2015relationship}:
\begin{equation}
\label{bte}
\frac{\partial f}{\partial t} + \frac{\boldsymbol{p}}{m} \cdot \nabla_{\boldsymbol{x}} f + F_{ex} \cdot \nabla_{\boldsymbol{p}} f = \left( \frac{\partial f}{\partial t} \right)_{coll},
\end{equation}
where the collision term captures changes due to pairwise particle interactions. For the hard-sphere model, collisions conserve momentum and kinetic energy, with post-collision velocities scattered randomly in the center-of-mass frame.

\subsection{Direct Simulation Monte Carlo (DSMC)}
The DSMC~\citep{bird1963approach} is a stochastic method for solving the BTE of \cref{bte} by simulating particle motion in a dilute gas, where each simulated particle represents $F_{N}$ physical molecules. The method divides space into small cells and evolves particle positions and velocities through three steps: (1) a deterministic drift step that moves particles along straight-line trajectories; (2) a deterministic wall collision step that resets particle velocities according to boundary conditions upon wall contact; and (3) a stochastic particle collision step, in which particle pairs within the same spatial cell collide according to a probability determined by the molecular interaction model. Further details on DSMC are provided in~\Cref{app:DSMC}.

% The DSMC~\citep{bird1963approach} is a stochastic method that simulates the particle motion to solve the BTE for a dilute gas. Unlike molecular dynamics, each particle here represents $F_{N}$ molecules in the physical system. It divides the space into small cells and evolves the position and velocity of particles in each cell. The evolution consists of three steps: 1) Drift, 2) Wall Collision, 3) Particle Collision.

% The first two steps are deterministic. The drift step moves the particles by assuming they move in straight lines without collision. The wall collision step checks if the particles collide with the wall and resets their velocity according to the boundary conditions. The last step is stochastic. The particles are sorted into spatial cells, and only particle pairs in the same cell are selected to collide. The collision probability depends on the molecular interaction model. For more detailed information about DSMC, please refer to \cref{app:DSMC}. 
\begin{figure*}[!t]
  \centering
  \begin{subfigure}[t]{0.99\textwidth}
    \centering
    \includegraphics[width=\linewidth]{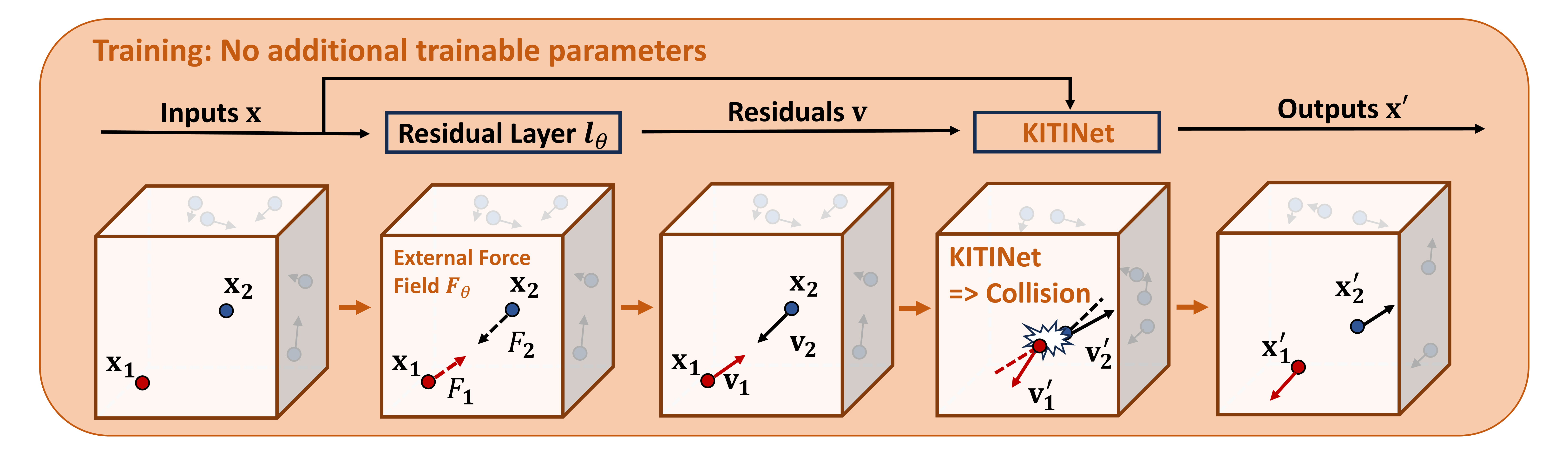}
    \caption{KITINet training process.}
    \label{fig:training}
  \end{subfigure}
  \hfill
  \begin{subfigure}[t]{0.99\textwidth}
    \centering
    \includegraphics[width=\linewidth]{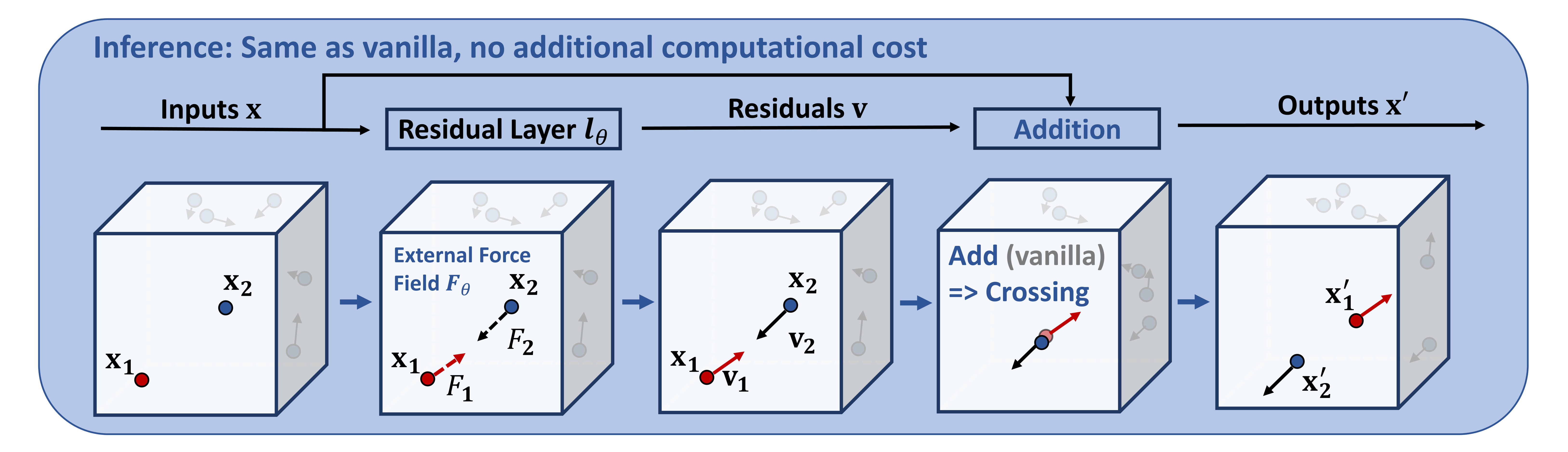}
    \caption{KITINet inference process.}
    \label{fig:inference}
  \end{subfigure}
  \caption{\rev{During training, KITINet replaces residual addition with an additional stochastic pairwise interaction with no additional trainable parameters. During evaluation, this interaction is disabled, and the update reduces exactly to $\boldsymbol{x}+\mathrm{d}t*\boldsymbol{v}$, so inference uses the original backbone architecture and incurs no additional inference cost. Only the parameters $\theta$ of the residual layer are learned.}}
  \label{fig:arch}
\end{figure*}

\subsection{Network Parameter Condensation}

\rev{Network parameter condensation~\citep{zhou2022towards,xu2025overview}
is a training phenomenon in which neuron weight vectors progressively
align and cluster around a small set of dominant directions. As training
proceeds, a nominally wide network develops a lower-dimensional parameter
organization, with many neurons concentrating around closely related
directions. Condensation therefore provides a geometric characterization
of the effective parameter complexity of an over-parameterized network.}

\rev{Prior work connects condensation to the embedding structure of
neural-network loss landscapes, reduced effective model complexity, and
improved generalization~\citep{zhang2021embedding,chen2024efficient}.
We quantify condensation using the average absolute cosine similarity
between neuron weight vectors and use it to characterize how KITINet
reshapes training dynamics. The formal definition is provided in
\cref{Sec: difinition of degree}.}
% For a weight matrix, the degree of condensation is quantified via the average absolute cosine similarity between neuron weight vectors: 
% $$\rho(\boldsymbol{W}) = \frac{1}{m(m-1)} \sum_{i \neq j} \frac{|\boldsymbol{w}_i^\top \boldsymbol{w}_j|}{\|\boldsymbol{w}_i\|_2 \|\boldsymbol{w}_j\|_2},$$
% where $\boldsymbol{w}_i$ denotes the incoming weight vector of neuron $i$ and $m$ is the layer width.
% Higher condensation (larger $\rho$) indicates that the effective number of distinct parameter directions is substantially smaller than the nominal width.

\section{Methodology: the KITINet Architecture}
\label{sec:kinetin_arch}

\rev{As illustrated in \cref{fig:arch}, KITINet augments a residual update during training with an additional operator with no additional trainable parameters. It reshapes feature groups as particles and residual outputs as velocity-like vectors, then applies a DSMC-inspired stochastic pairwise interaction before completing the update. During inference, KITINet is disabled, and the layer reduces exactly to the original residual addition $\boldsymbol{x}+\mathrm{d}t\,\boldsymbol{v}$, thereby incurring no additional inference cost.}\rev{For a layer with input $\boldsymbol{x}\in\mathbb{R}^D$ and residual output $\boldsymbol{v}\in\mathbb{R}^D$, we choose a collision-head dimension $H$ and reshape them as $\boldsymbol{X},\boldsymbol{V}\in\mathbb{R}^{H\times N}$, where $N=D/H$. This defines $N$ feature groups in an $H$-dimensional interaction space. The $i$-th \emph{column} of $\boldsymbol{X}$ and $\boldsymbol{V}$, denoted $\boldsymbol{x}_i,\boldsymbol{v}_i\in\mathbb{R}^{H}$, represents the position-like and velocity-like vectors of group $i$.}

Analogous to multi-head attention in Transformers, which employs multiple heads to capture diverse feature dependencies, our collision head structure defines the dimensionality of the space in which particles interact. However, our method diverges from Transformers in its spatial organization: whereas Transformers partition features into independent representation subspaces, we unify all particles within a single shared $H$-dimensional space to facilitate collective inter-particle collisions.

\begin{algorithm}[tb!]
\caption{KITINet (with training and inference).}
\label{alg:KITINet_v}
\begin{algorithmic}[1]
\STATE \textbf{Input:} Layer input $\boldsymbol{x}\in \mathbb{R}^D$, residual $\boldsymbol{v}\in \mathbb{R}^D$, time step $\mathrm{d}t$, number of collision heads $H$, and coll\_coef;
\STATE \textbf{Output:} Layer output $\boldsymbol{x'}\in \mathbb{R}^D$.
\STATE If model is in the inference mode, \textbf{Return} $\boldsymbol{x} + \mathrm{d}t*\boldsymbol{v}$;
\STATE Reshape $\boldsymbol{x,v}$ to $\boldsymbol{X,V} \in \mathbb{R}^{H \times N}$, where $N = {D}/{H}$;
\STATE Calculate relative properties $X_r, V_r$ and center-of-mass properties $\boldsymbol{X}_{cm}, \boldsymbol{V}_{cm}$ by \cref{eq: xv_r}.
%\STATE Get relative receding velocity $\boldsymbol{V}_r^{\text{receding}}$ from $V_r$ by Alg.~\ref{alg:RelativeRecedingVelocity}
\STATE Calculate the full velocity change $\Delta \boldsymbol{V}$ by \cref{eq: delta_v};
\STATE Select collision pairs by \cref{eq: coll_mask};
\STATE Apply velocity and position change by \cref{eq: xv_new}, get new position $\boldsymbol{X'}$;
\STATE \textbf{Return} $\boldsymbol{x'}$ flattened from $\boldsymbol{X'}$;
\end{algorithmic}
\end{algorithm}
Specifically, our KITINet simulates the collisions by imitating the DSMC method:
\begin{enumerate}
%\item (Optional) Calculate the velocity before collision if a-edition KITINet is chosen:
%$$\boldsymbol {v} _{i} = \boldsymbol{v_0} _{i} + \boldsymbol {a} _{i}dt.$$
\item Compute relative distance, relative velocity, center-of-mass position and velocity for $N$ particles:
\begin{equation}
\begin{aligned}
(X_r)_{i,j}&=|\boldsymbol {x} _{i}-\boldsymbol {x} _{j}|, \quad
(V_r)_{i,j}=|\boldsymbol {v} _{i}-\boldsymbol {v} _{j}|\\
(\boldsymbol{X}_{cm})_{i,j}&=\frac{1}{2}(\boldsymbol {x} _{i}+\boldsymbol {x} _{j}),\quad(\boldsymbol{V}_{cm})_{i,j}=\frac{1}{2}(\boldsymbol {v} _{i}+\boldsymbol {v} _{j}).
\end{aligned}
\label{eq: xv_r}
\end{equation}
    ($(X_r)_{i,j}$, $(V_r)_{i,j}$ are scalars; $(\boldsymbol{X}_{cm})_{i,j}$, $(\boldsymbol{V}_{cm})_{i,j}$ are vectors.)

\item Compute velocity change $\Delta \boldsymbol{V}$:
\begin{equation}
\begin{aligned}
(\Delta \boldsymbol{V})_{i,j}
&= (\boldsymbol{V}_{cm})_{i,j} + \frac{1}{2}(V_r)_{i,j} \boldsymbol {n}_{i,j} - \boldsymbol {v}_{i}, \\
(\Delta \boldsymbol{V})_{j,i}
&= (\boldsymbol{V}_{cm})_{j,i} + \frac{1}{2}(V_r)_{j,i} \boldsymbol {n}_{j,i} - \boldsymbol {v}_{j}.
\end{aligned}
\label{eq: delta_v}
\end{equation}
    where $\boldsymbol {n}_{i,j}$ is uniformly random on the $H$-dim unit sphere, $\boldsymbol {n}_{j,i} =- \boldsymbol {n}_{i,j}$, adapted from~\cref{eq:DSMC_deltav,eq: DSMC_vr}.

\item Accept collision for pair $i,j$ if
\begin{equation}
    \frac{(V_r)_{i,j} \cdot (U_r)_{i,j}}{v_r^{max}}>1-\text{coll}\_{\textrm{coef}},
    \label{eq: coll_mask}
\end{equation}
    \rev{where $(U_r)_{i,j}=e^{-(X_r)_{i,j}}$ and $v_r^{max}=\max(V_r)$. The exponential term is a dimensionless, distance-dependent interaction kernel. Unlike the cell-restricted acceptance rule in DSMC, KITINet permits all pairs and uses $\mathrm{coll\_coef}$ as a tunable threshold.}

\item Update velocity and position via $\boldsymbol {x} _{i}' = \boldsymbol {x}^{*} _{i} + \mathrm{d}t*\boldsymbol {v} _{i}'$:
\begin{equation}
\begin{aligned}
\boldsymbol {v} _{i}' &=\boldsymbol {v} _{i} + \sum_{j \in \mathcal{C}_i} (\Delta \boldsymbol{V})_{i,j}, \\
\boldsymbol {x}^{*} _{i} &= \frac{1}{1+k}\Big(\boldsymbol {x} _{i} + \sum_{j \in \mathcal{C}_i} (\boldsymbol{X}_{cm})_{i,j}\Big),
\end{aligned}
\label{eq: xv_new}
\end{equation}
\rev{where $\mathcal{C}_i$ is the set of accepted partners of particle $i$ and $k=|\mathcal{C}_i|$. The averaged position $\boldsymbol{x}^{*}_i$ accumulates the accepted pairwise interactions within one residual update; this differs from the collision step in classical DSMC and is evaluated in \cref{subsubsec:Ablation}.}
%\item (Optional) Record the velocity after collision for the next layer if a-edition KITINet is chosen:
%$$\boldsymbol{v_0} _{i}=\boldsymbol {v} _{i}^{*}.$$
\end{enumerate}

\rev{The method is summarized in \cref{alg:KITINet_v}. The cited stochastic particle schemes obtain BTE approximations under three structural assumptions: (i) spatial homogeneity; (ii) exchangeable particles with molecular-chaos sampling of pre-collision pairs; and (iii) binary elastic collisions that conserve pairwise momentum and kinetic energy~\citep{Karabulut_2007,doi:10.1143/JPSJ.49.2042,Babovsky1986SimulationScheme}. KITINet follows the corresponding pairwise construction: every feature group is processed by the same rule based on the relative quantities $(X_r)_{i,j}$ and $(V_r)_{i,j}$, while pair scoring and random scattering are symmetric in $i$ and $j$ and use their pre-interaction states. For each accepted binary interaction, the update in \cref{eq: delta_v} preserves the center-of-mass velocity and relative-speed norm, thereby conserving the momentum and kinetic energy of that pair. All-pairs eligibility, the exponential distance kernel, the tunable threshold, and averaged position updates determine how these pairwise interactions are selected and incorporated into the neural update. The module has no additional trainable parameters, and its training-time complexity is $O(N^{2}H)=O(D^2/H)$.}

\section{\texorpdfstring{\rev{Mechanistic Interpretation and Condensation Analysis}}{Mechanistic Interpretation and Condensation Analysis}}
\label{sec:mech}
\rev{We analyze the connection between KITINet's structured stochastic interactions and parameter condensation from complementary physical and theoretical perspectives. The physical analogy motivates the mechanism, while the surrogate theorem analyzes analogous condensation-like dynamics in an idealized simplified setting. The corresponding empirical measurements are presented separately in \cref{sec:Analysis}.}

\subsection{Physics Analysis: From Collisions to Parameter Condensation}
\label{subsec:physics analysis}
In statistical mechanics, the viscosity of a fluid measures its resistance to deformation and governs the rate of entropy production. For a system of hard-sphere particles, the microscopic stress flux $J$ decomposes into kinetic and collisional contributions~\citep{Wainwright1964}:
\begin{equation}
J(t) = m \sum_{i=1}^{N} v_i^2(t) + \frac{1}{2} \sum_{i=1}^{N} \sum_{j=1}^{N} |\boldsymbol F_{coll,ij} \times (\boldsymbol{x_i} - \boldsymbol{x_j})|,
\label{eq: J}
\end{equation}
where $\boldsymbol F_{coll,ij}$ is the force exerted on particle $i$ during a hard-sphere collision with particle $j$. Via the Green-Kubo relations~\citep{10.1063/1.1740082,doi:10.1143/JPSJ.12.570}, the macroscopic viscosity $\eta$ is the time integral of the stress autocorrelation:
\begin{equation}
\eta = \frac{1}{VkT} \int _{0} ^{\infty} \langle J(t)J(0) \rangle \, dt = \eta^{K} + \eta^{C},
\label{eq: GK-eta}
\end{equation}
where the kinetic term $\eta^{K}$ is the Chapman-Enskog viscosity, $\eta^{C}$ is the collision-induced correction, and the cross term vanishes for the DSMC hard-sphere model~\citep{Garcia2025_DSMC}.

\rev{Thus, collision-mediated momentum transfer contributes an additional collisional component to the macroscopic viscosity and promotes the relaxation of heterogeneous particle velocities. In KITINet, the analogous pairwise interactions redistribute residual updates across feature groups, reducing local disparities in their directions and magnitudes. This structured redistribution produces a smoother, more coordinated update field, enabling the input--output mapping to be learned with less heterogeneous parameter adjustment and thereby promoting parameter condensation.}

% \textbf{From viscosity to smoothness.} Higher viscosity implies stronger resistance to rapid variations in the velocity field. In the neural network analogue, the \emph{velocity field} is the layer output $\boldsymbol{v}$ and the \emph{external force} is the gradient signal driving parameter updates. By enhancing the effective viscosity, collisions attenuate high-frequency fluctuations in $\boldsymbol{v}$, yielding a smoother force field. The network is therefore relieved from producing sharp, brittle feature transformations.

% \textbf{From smoothness to condensation.} A smoother layer-wise mapping means neighboring neurons receive more coherent gradient signals. Under the embedding principle of loss landscapes~\citep{Zhang2021EmbeddingPO}, gradient descent naturally converges toward low-complexity critical points. The collision-enhanced viscosity accelerates this process by suppressing gradient variance across neurons within the same layer, inducing their weight vectors to align. The causal chain is thus: \textbf{collisions} $\rightarrow$ \textbf{enhanced viscosity} $\rightarrow$ \textbf{smoother force field} $\rightarrow$ \textbf{coherent within-layer gradients} $\rightarrow$ \textbf{accelerated condensation}. Empirical validation is provided in~\cref{sec:Analysis}.

\subsection{Theoretical Analysis in a Simplified Setting}
\label{subsec:theory analysis}
\rev{We next analyze a tractable model that isolates the effect of collision-like perturbations. The model is a two-layer over-parameterized linear regression network in which collision effects are represented by uniform stochastic perturbations to the second-layer weights under an idealized thermal-equilibrium assumption. This formulation isolates the weight-norm and alignment dynamics induced by collision-like perturbations. \cref{main theorem} therefore characterizes these dynamics; the detailed setup and proof are provided in \cref{Appendix: theory}.}

\begin{theorem}
  \label{main theorem}
     Under the setting in~\cref{theory setting} and assuming the simplified thermal-equilibrium perturbation model, the introduction of collision-like stochasticity transforms the convergence dynamics from single-phase exponential convergence to a two-phase process: \\
\textbf{(1) Weight-norm decay:} The norms of the neuron weight vectors first decay to a small scale, inducing rapid reorientation in the low-weight regime. \\
\textbf{(2) Condensation:} The model subsequently converges to a sparse solution through condensation-like alignment dynamics.
  \end{theorem}
  \textbf{Proof Sketch.} Without collisions,~\cref{thm: case1} proves exponential convergence to a dense solution. With the collision model, the iterative process becomes a Markov chain. \cref{algorithm 2 decay} proves that at each step, the expected weight norms strictly decrease until they reach a small scale, after which known results on gradient dynamics in the low-weight regime~\citep{Maennel2018GradientDQ, Lyu2021GradientDO, zhou2022towards} imply condensation.
  
  \rev{\textbf{Insights.} Under the simplified perturbation model, the analysis yields a two-phase process consisting of weight-norm decay followed by condensation-like alignment. By isolating collision-induced weight dynamics with idealized stochastic perturbations, it provides a theoretical mechanism for the condensation observed empirically.}

\section{Experiments}
\label{sec: Experiments}

\rev{We first evaluate KITINet across four NLP settings in \cref{subsec:nlp}, then test canonical residual vision models on CIFAR in \cref{subsec:image} and neural operators for PDE solving in \cref{subsec:pde solving}. We next conduct controlled studies with FNO, including comparisons with alternative stochastic training strategies, an ablation of the position update, and hyperparameter-sensitivity analyses in \cref{subsec:controlled_studies}. Finally, in \cref{sec:Analysis}, we examine parameter condensation in synthetic networks and selected ResNet and FNO layers before discussing the method's computational and empirical limitations.}

\subsection{Natural Language Processing Tasks}
\label{subsec:nlp}
\rev{To evaluate KITINet across natural language processing regimes, we consider LLM pre-training from scratch in \cref{subsubsec:pretrain}, continued pre-training for domain adaptation in \cref{subsubsec:continued_pretrain}, parameter-efficient supervised fine-tuning for chemical capability transfer in \cref{subsubsec:chem_sft}, and encoder fine-tuning for text classification in \cref{subsubsec:bert}.}

\subsubsection{LLM Pre-training from Scratch}
\label{subsubsec:pretrain}

\begin{table}[b]
\centering
\caption{Accuracy(\%)$\uparrow$ of vanilla and KITINet-plugin GPT2. Values are reported as mean $\pm$ standard error.}
\label{tab:gpt2-acc}
\scriptsize  % 改用更小字号
\setlength{\tabcolsep}{1.5pt}
\renewcommand\arraystretch{0.75}
\resizebox{\linewidth}{!}{%
\begin{tabular}{lccccc}
\toprule 
Model & MMLU & ARC\_C & ARC\_E & HellaSwag & WinoGrande \\ 
\midrule
GPT2~\citep{gpt2} & 24.9 $\pm$ 0.35 & 21.8 $\pm$ 1.11 & 43.3 $\pm$ 1.01 & 38.4 $\pm$ 0.49 & 50.9 $\pm$ 1.41 \\
KITI-GPT2 & \textbf{25.1 $\boldsymbol{\pm}$ 0.35} & \textbf{22.6 $\boldsymbol{\pm}$ 1.11} & \textbf{43.5 $\boldsymbol{\pm}$ 1.02} & \textbf{38.8 $\boldsymbol{\pm}$ 0.48} & \textbf{51.2 $\boldsymbol{\pm}$ 1.40} \\
\midrule
GPT2-medium~\citep{gpt2} & \textbf{27.0 $\boldsymbol{\pm}$ 0.37} & 26.5 $\pm$ 1.12 & 52.3 $\pm$ 1.02 & 46.0 $\pm$ 0.49 & 53.4 $\pm$ 1.40 \\
KITI-GPT2-med & 26.2 $\pm$ 0.36 & \textbf{27.6 $\boldsymbol{\pm}$ 1.13} & \textbf{52.5 $\boldsymbol{\pm}$ 1.02} & \textbf{46.3 $\boldsymbol{\pm}$ 0.50} & \textbf{53.8 $\boldsymbol{\pm}$ 1.40} \\
\midrule
GPT2-large~\citep{gpt2} & 25.9 $\pm$ 0.35 & 28.5 $\pm$ 1.29 & 57.5 $\pm$ 1.02 & 46.7 $\pm$ 0.50 & 55.2 $\pm$ 1.39 \\
KITI-GPT2-large & \textbf{26.1 $\boldsymbol{\pm}$ 0.36} & \textbf{28.6 $\boldsymbol{\pm}$ 1.30} & \textbf{57.7 $\boldsymbol{\pm}$ 1.01} & \textbf{47.2 $\boldsymbol{\pm}$ 0.50} & \textbf{55.7 $\boldsymbol{\pm}$ 1.40} \\
\midrule
GPT2-xl~\citep{gpt2} & 26.6 $\pm$ 0.36 & 31.8 $\pm$ 1.31 & 62.2 $\pm$ 0.99 & 50.6 $\pm$ 0.51 & 58.2 $\pm$ 1.38 \\
KITI-GPT2-xl & \textbf{27.2 $\boldsymbol{\pm}$ 0.37} & \textbf{31.9 $\boldsymbol{\pm}$ 1.32} & \textbf{62.9 $\boldsymbol{\pm}$ 0.99} & \textbf{51.0 $\boldsymbol{\pm}$ 0.51} & \textbf{58.5 $\boldsymbol{\pm}$ 1.39} \\
\bottomrule
\end{tabular}%
}
\end{table}

\textbf{Dataset \& models}. We pre-train the GPT-2 series from 0.1B to 1.5B parameter models with the standard next token prediction loss. We replace the residual connection after the attention module with a KITINet layer, obtaining a new series of models named KITINet-GPT-2.  Our training corpus is a 30B token high-quality composition of web text (FineWeb-edu \citep{lozhkov2024fineweb-edu}), mathematics (MegaMath \citep{zhou2025megamath}), and code (OpenCoder \citep{Huang2024OpenCoderTO}), which reflects current state-of-the-art data curation practices. The evaluation of GPT-2 and KITINet-GPT-2 is based on a diverse set of challenging downstream benchmarks, including knowledge-intensive tasks (MMLU \citep{hendryckstest2021}, ARC \citep{allenaiarc}) and commonsense reasoning tasks (HellaSwag \citep{zellers2019hellaswag}, WinoGrande \citep{ai2winogrande}).

\rev{\textbf{Results.} Table~\ref{tab:gpt2-acc} reports accuracy with standard errors. KITINet improves 18 of the 20 model--benchmark comparisons, with an average gain of approximately $0.5$ percentage points. The gains span every model scale and benchmark, demonstrating a broad positive trend across the evaluation suite.}

% \textbf{Training cost analysis.} The KITINet module introduces $O(D^2/H)$ additional computation per residual connection during training. For the GPT-2 series, this translates to a per-step FLOPs overhead of approximately $4\%$ (GPT-2), $6\%$ (GPT-2-medium), $7\%$ (GPT-2-large), and $8\%$ (GPT-2-xl). However, KITINet consistently reaches target accuracy levels with approximately $20\%$ fewer training steps, yielding a \emph{net wall-clock time reduction} of $12$--$16\%$ across all model sizes. The training overhead is thus more than offset by faster convergence. Memory overhead is negligible since KITINet introduces no trainable parameters and intermediate pairwise tensors are discarded after each forward pass.

\subsubsection{\texorpdfstring{\rev{LLM Continued Pre-training for Domain Adaptation}}{LLM Continued Pre-training for Domain Adaptation}}
\label{subsubsec:continued_pretrain}
\begin{table}[t]
\centering
\caption{Perplexity$\downarrow$ of vanilla and KITINet GPT2.}
\label{tab:gpt-perplexity}
\renewcommand\arraystretch{0.92}
\setlength{\tabcolsep}{3pt}
\resizebox{\linewidth}{!}{%  % 关键: \columnwidth 改为 \linewidth
\begin{tabular}{lcccc}
\toprule 
Model & GSM8K & MATH & MATHQA & OCW\\ 
\midrule
GPT2~\citep{gpt2} & 17.34 & 12.35 & 31.24 & 7.34 \\
KITI-GPT2 & \textbf{17.29} & \textbf{12.18} & \textbf{30.39} & \textbf{7.21} \\
\midrule
GPT2-medium~\citep{gpt2} & \textbf{13.72} & 11.18 & 23.68 & 6.35 \\
KITI-GPT2-med & 13.82 & \textbf{11.02} & \textbf{23.05} & \textbf{6.21} \\
\midrule
GPT2-large~\citep{gpt2} & 11.59 & 10.27 & 19.56 & 5.53 \\
KITI-GPT2-large & \textbf{11.20} & \textbf{10.06} & \textbf{18.81} & \textbf{5.40} \\
\midrule
GPT2-xl~\citep{gpt2} & 10.59 & 9.78 & 17.48 & 5.23 \\
KITI-GPT2-xl & \textbf{10.42} & \textbf{9.63} & \textbf{16.99} & \textbf{5.10} \\
\bottomrule
\end{tabular}%
}
\end{table}

% \begin{table}[t]
% \centering
% \caption{Perplexity of vanilla and KITINet GPT2.}
% \label{tab:gpt-perplexity}
% 

% \resizebox{0.95\columnwidth}{!}{%
% \begin{tabular}{lcccc}
% \toprule 
% Model & GSM8K $\downarrow$ & MATH $\downarrow$ & MATHQA $\downarrow$ & OCW $\downarrow$\\ 
% \midrule
% GPT2~\citep{gpt2} & 17.34 & 12.35 & 31.24 & 7.34 \\
% KITI-GPT2 & \textbf{17.29} & \textbf{12.18} & \textbf{30.39} & \textbf{7.21} \\
% \midrule
% GPT2-medium~\citep{gpt2} & \textbf{13.72} & 11.18 & 23.68 & 6.35 \\
% KITI-GPT2-med & 13.82 & \textbf{11.02} & \textbf{23.05} & \textbf{6.21} \\
% \midrule
% GPT2-large~\citep{gpt2} & 11.59 & 10.27 & 19.56 & 5.53 \\
% KITI-GPT2-large & \textbf{11.20} & \textbf{10.06} & \textbf{18.81} & \textbf{5.40} \\
% \midrule
% GPT2-xl~\citep{gpt2} & 10.59 & 9.78 & 17.48 & 5.23 \\
% KITI-GPT2-xl & \textbf{10.42} & \textbf{9.63} & \textbf{16.99} & \textbf{5.10} \\
% \bottomrule
% \end{tabular}%
% }
% 
% \end{table}

\rev{We additionally continue pre-training GPT-2 on Open-Web-Math~\citep{paster2023openwebmath}. Starting from the official GPT-2 weights, matched baseline and KITINet variants are trained for 30 billion tokens and evaluated on GSM8K~\citep{cobbe2021gsm8k}, MATH~\citep{hendrycksmath2021}, MATH\_QA~\citep{amini-etal-2019-mathqa}, and OCW~\citep{lewkowycz2022solving} using the perplexity of correct answers. As shown in \cref{tab:gpt-perplexity}, KITINet has lower perplexity in 15 of the 16 model--benchmark comparisons; the GPT-2-medium result on GSM8K is the exception.}

\subsubsection{\texorpdfstring{\rev{LLM Supervised Fine-tuning with LoRA}}{LLM Supervised Fine-tuning with LoRA}}
\label{subsubsec:chem_sft}

\begin{table}[b]
\centering
\caption{\rev{Qwen-3.5 chemical capability SFT evaluated on the nine-task ChemBench suite and end-to-end training time. Values are reported as mean $\pm$ standard error.}}\label{tab:chem_sft}
\setlength{\tabcolsep}{1pt}\renewcommand\arraystretch{0.92}{\color{black}\begin{tabular*}{\linewidth}{@{\extracolsep{\fill}}llccc@{}}
\toprule Scale & Adapter & \shortstack{Task-Macro\\Accuracy(\%)$\uparrow$} & \shortstack{Sample\\Accuracy (\%)$\uparrow$} & \shortstack{Training\\Time (s)} \\
\midrule
\multirow{2}{*}{0.8B} & LoRA~\citep{hu2022lora} & 27.65 $\pm$ 0.40 & 27.87 $\pm$ 0.35 & 12{,}038.8 \\
& KITI-LoRA & \textbf{30.48 $\boldsymbol{\pm}$ 0.41} & \textbf{30.79 $\boldsymbol{\pm}$ 0.36} & 12{,}768.5 \\
\midrule
\multirow{2}{*}{2B} & LoRA~\citep{hu2022lora} & 36.99 $\pm$ 0.41 & 35.73 $\pm$ 0.36 & 18{,}199.1 \\
& KITI-LoRA & \textbf{38.88 $\boldsymbol{\pm}$ 0.41} & \textbf{38.67 $\boldsymbol{\pm}$ 0.37} & 19{,}135.8 \\
\bottomrule
\end{tabular*}}
\end{table}

\rev{\textbf{Dataset \& models}. To evaluate domain-specific capability transfer under parameter-efficient supervised fine-tuning, we fine-tune Qwen3.5-Base models~\citep{qwen2026qwen35} at the 0.8B and 2B scales on 853,657 chemical instruction records derived from ChemPile~\citep{mirza2025chempile}. We compare standard LoRA SFT with a scale-matched LoRA + KITINet treatment on the nine-task ChemBench suite introduced with ChemLLM~\citep{zhang2024chemllm}. Each item is scored by choice log-probability under a fixed few-shot prompt. We report task-macro accuracy, defined as the unweighted mean of the nine task accuracies on a 0--1 scale, and sample accuracy, defined as the percentage of correct predictions over all 4,008 evaluation records.}

\rev{\textbf{Settings.} All experiments start from Qwen3.5 Base checkpoints rather than instruction-tuned checkpoints. We fine-tune the models with LoRA~\citep{hu2022lora} using rank 8, scaling factor $\alpha=32$, dropout 0.05, and all language-layer linear suffixes as target modules. Training uses no quantization, BF16 precision with TF32 enabled, a sequence cutoff of 2,048 tokens, and AdamW~\citep{loshchilov2019decoupled} with a learning rate of $5\times10^{-5}$, cosine decay, and a warmup ratio of 0.03. Each model is trained for one epoch with gradient accumulation of 1 and gradient checkpointing enabled. At 0.8B, each treatment uses four H200 GPUs with a per-device batch size of 8; at 2B, each treatment uses two H200 GPUs with a per-device batch size of 16. Both settings therefore have an effective batch size of 32. Within each matched pair, the baseline and KITINet runs use the same base checkpoint, LoRA configuration, training examples and order, optimizer schedule, effective batch size, and training seed. KITINet is applied at three MLP residual connections only during training and is disabled at evaluation, leaving the standard LoRA inference graph unchanged.}

\rev{\textbf{Results.} As shown in \cref{tab:chem_sft}, KITINet raises task-macro accuracy from $0.2765$ to $0.3048$ at 0.8B and from $0.3699$ to $0.3888$ on average at 2B, corresponding to gains of $0.0282$ and $0.0189$, respectively. Sample accuracy improves by $2.92$ and $2.94$ percentage points, and both 2B training seeds favor KITINet. The Time column reports the end-to-end \texttt{train\_runtime} recorded by the Trainer. Within each scale and seed, the baseline and KITINet runs use the same GPU allocation, effective batch size, training data and order, and number of optimizer steps; their paired differences therefore estimate KITINet's runtime overhead under the evaluated setup. The same table shows that training time increases by only $6.06\%$ at 0.8B and $5.15\%$ on average at 2B. Since KITINet is disabled at evaluation, it adds no inference-time overhead and leaves the standard LoRA inference graph unchanged.}

\subsubsection{\texorpdfstring{\rev{Encoder Fine-tuning for Text Classification}}{Encoder Fine-tuning for Text Classification}}
\label{subsubsec:bert}
% \begin{table}[t]
% \centering
% \caption{Accuracy comparisons on IMDb and SNLI.}
% \label{Tab: txt_exp}
% 

% \renewcommand\arraystretch{1.0}
% \resizebox{0.95\columnwidth}{!}{%
% \begin{tabular}{lcc}
% \toprule 
% Model & IMDb $\uparrow$ & SNLI $\uparrow$ \\
% \midrule
% BERT-cased \citep{Devlin2019BERTPO} & 91.45\% & 89.28\%  \\
% KITI-BERT-cased & \textbf{92.96\%} & \textbf{90.26\%} \\
% \midrule
% BERT-uncased \citep{Devlin2019BERTPO} & 93.42\% & 89.02\% \\
% KITI-BERT-uncased & \textbf{94.53\%} & \textbf{90.56\%} \\
% \bottomrule
% \end{tabular}%
% }
% 
% \end{table}

\begin{table}[b]
\centering
\caption{Accuracy(\%)$\uparrow$ comparisons on IMDb and SNLI.}
\label{Tab: txt_exp}
\fontsize{10pt}{12pt}\selectfont
\renewcommand\arraystretch{0.75}\begin{tabular*}{0.8\linewidth}{@{\extracolsep{\fill}}lcc@{}}
\toprule 
Model & IMDb & SNLI \\
\midrule
BERT-cased \citep{Devlin2019BERTPO} & 91.45 & 89.28  \\
KITI-BERT-cased & \textbf{92.96} & \textbf{90.26} \\
\midrule
BERT-uncased \citep{Devlin2019BERTPO} & 93.42 & 89.02 \\
KITI-BERT-uncased & \textbf{94.53} & \textbf{90.56} \\
\bottomrule
\end{tabular*}
\end{table}

\textbf{Dataset \& models.} We use two benchmark datasets: IMDb \citep{maas-EtAl:2011:ACL-HLT2011} for sentiment classification, and SNLI \citep{bowman-etal-2015-large} for natural language inference, both of which are standard benchmarks for natural language understanding and reasoning. We adopt BERT \citep{Devlin2019BERTPO} as the backbone and \rev{integrate KITINet into its final Transformer layer (KITI-BERT) without adding trainable parameters. Because KITINet is disabled at inference, it incurs no additional inference cost.}

\textbf{Settings.}  
We do experiments with two pre-trained model variants, including bert-base-cased and bert-base-uncased. We set the tokenizer corresponding to the pre-trained model to process input tokens. Both KITI-BERT and BERT are trained by Adam with a batch size of 32 and the same random seed. We set the learning rate to $2\times10^{-5}$ and the total fine-tuning training epochs to $40$.

\rev{\textbf{Results.} As shown in \cref{Tab: txt_exp}, KITI-BERT consistently outperforms its matched BERT baseline across all four comparisons. For BERT-cased, KITINet raises accuracy from $91.45\%$ to $92.96\%$ on IMDb and from $89.28\%$ to $90.26\%$ on SNLI, corresponding to gains of $1.51$ and $0.98$ percentage points. For BERT-uncased, the corresponding accuracies increase from $93.42\%$ to $94.53\%$ on IMDb and from $89.02\%$ to $90.56\%$ on SNLI, yielding gains of $1.11$ and $1.54$ percentage points.}

\subsection{Image Classification on CIFAR}
\label{subsec:image}
\textbf{Dataset and models}. We conduct our evaluations on CIFAR with 50K training images and 10K test images. Our training setup follows \citep{he2016deep}, including models in varying depth (ResNet-34, ResNet-50, ResNet-101, and ResNet-152). To balance performance and cost, we only integrate our module in the last stage of the ResNet architecture.

\textbf{Settings}. The models are trained by SGD with batch size 128, momentum coefficient 0.9, and weight decay $5\times10^{-4}$. The learning rate is initialized to $1$ for the quadratic integration matrix in the implementation of Dit-ResNet~\citep{liu2024dendritic} and $0.1$ for all other parameters and is decayed by a factor of ten at the 80th and 120th epochs, completing training after 160 epochs. We apply standard augmentation to images: padding with 4 pixels on each side, followed by a random $32\times32$ crop, and random horizontal flipping. For evaluation, we use the original $32\times32$ resolution without augmentation. Following \citep{liu2024dendritic}, we incorporate quadratic neurons specifically into the same layer.

\begin{table}[b]  % r=右侧, 稍缩小宽度留出间距
\centering
\caption{Accuracy(\%)$\uparrow$ of KITINet and ResNet-based models. CIFAR-100 values are reported as mean $\pm$ standard error.}
\label{Tab: img_exp}
\renewcommand\arraystretch{0.92}
\resizebox{\linewidth}{!}{%  % 改用\linewidth, 即0.48\textwidth
\begin{tabular}{lccc}
\toprule 
Model & CIFAR10 & CIFAR100 & MFLOPs\\ 
\midrule
ResNet-34~\citep{he2016deep} & 94.48 & 77.97 $\pm$ 0.12 & 73.5\\
Dit-ResNet-34~\citep{liu2024dendritic} & 94.45 & 78.14 $\pm$ 0.07 & 73.5\\
KITI-ResNet-34 & \textbf{95.04} & \textbf{78.67 $\boldsymbol{\pm}$ 0.10}& 73.6\\
\midrule
ResNet-50~\citep{he2016deep} & 94.75 & 78.27 $\pm$ 0.09 & 83.7\\
Dit-ResNet-50~\citep{liu2024dendritic} & 94.53 & 78.61 $\pm$ 0.05 & 83.7\\
KITI-ResNet-50 & \textbf{95.18} & \textbf{78.75 $\boldsymbol{\pm}$ 0.10} & 85.9\\
\midrule
ResNet-101~\citep{he2016deep} & 94.71 & 78.39 $\pm$ 0.08 & 159.2\\
Dit-ResNet-101~\citep{liu2024dendritic} & 94.98 & 78.88 $\pm$ 0.05 & 159.2\\
KITI-ResNet-101 & \textbf{95.01} & \textbf{79.09 $\boldsymbol{\pm}$ 0.03} & 161.3\\
\midrule
ResNet-152~\citep{he2016deep} & 94.67 & 78.41 $\pm$ 0.07 & 234.7\\
Dit-ResNet-152~\citep{liu2024dendritic} & 95.21 & 78.84 $\pm$ 0.04 & 234.7\\
KITI-ResNet-152 & \textbf{95.67} & \textbf{79.48 $\boldsymbol{\pm}$ 0.03} & 236.8\\
\bottomrule
\end{tabular}%
}
\end{table}

\rev{\textbf{Results.} \cref{Tab: img_exp} compares KITINet with vanilla ResNet~\citep{he2016deep} and Dit-ResNet~\citep{liu2024dendritic}. KITINet obtains the highest reported accuracy in each depth--dataset configuration without adding trainable parameters and incurs no additional inference cost. Averaged over the four depths, the gains over vanilla ResNet are $0.29$ percentage points on CIFAR-10 and $0.65$ on CIFAR-100. Together, these results demonstrate consistent improvements across all evaluated ResNet depths and both datasets.}
% Moreover, the FLOPs overhead is modest as shown in the right column.
% $+0.1\%$ to $+0.9\%$ for the configurations shown in~\cref{Tab: img_exp}. This is because KITINet is applied only to the last stage of the ResNet. The training wall-clock time increases by less than $2\%$ for ResNet-34 through ResNet-152. We note that the standard errors on CIFAR-100 (reported in parentheses) indicate that the gains over Dit-ResNet for ResNet-50 and ResNet-101 are near the margin of statistical significance; the direction of improvement is consistent but the magnitude is modest.

\subsection{Learning Neural Operator for PDE-solving}
\label{subsec:pde solving}

% \begin{table}[t]
% \centering
% \caption{Performance of vanilla and KITINet models on PDE-solving tasks.}
% \label{Tab:pde_exp}
% 
% \renewcommand\arraystretch{0.95}

% \resizebox{0.95\columnwidth}{!}{%
% \begin{tabular}{lcc}
% \toprule
% Problem & Model & Relative MSE $\downarrow$ \\  
% \midrule
% \multirow{2}{*}{Burgers' Equation} 
%  & FNO~\citep{li2021fno} & 0.00217 \\
%  & KITI-FNO & \fst{0.00166} \\
% \midrule
% \multirow{2}{*}{NS Equation}
%  & FNO~\citep{li2021fno} & 0.12023 \\
%  & KITI-FNO & \fst{0.11346} \\
% \midrule
% \multirow{2}{*}{Heat Equation}
%  & FNO~\citep{li2021fno} & 0.07054 \\
%  & KITI-FNO & \fst{0.05113} \\
% \midrule
% \multirow{2}{*}{Airfoil$^{*}$}
%  & OFormer~\citep{li2023oformer} & 16.39461 \\
%  & KITI-OFormer & \fst{15.49034} \\
% \bottomrule
% \end{tabular}%
% }

% % 
% \footnotesize
% $^{*}$ Airfoil problem uses the Relative Root MSE measurement.
% \end{table}

\begin{table}[b]  % r=右侧, 宽度占48%行宽
\centering
\caption{Performance of vanilla and KITINet models on PDE-solving tasks. \rev{Values are reported as mean $\pm$ standard error.}}
\label{Tab:pde_exp}
\renewcommand\arraystretch{0.95}
\resizebox{0.95\linewidth}{!}{%  % 注意用\linewidth而非\columnwidth
\begin{tabular}{lcc}
\toprule
Problem & Model & Relative MSE$\downarrow$ \\  
\midrule
\multirow{2}{*}{Burgers' Equation} 
 & FNO~\citep{li2021fno} & 0.00217 $\pm$ 0.00008 \\
 & KITI-FNO & \textbf{0.00166 $\boldsymbol{\pm}$ 0.00005} \\
\midrule
\multirow{2}{*}{NS Equation}
 & FNO~\citep{li2021fno} & 0.12023 $\pm$ 0.00587 \\
 & FNO + KITINet & \textbf{0.11346 $\boldsymbol{\pm}$ 0.00717} \\
\midrule
\multirow{2}{*}{Heat Equation}
 & FNO~\citep{li2021fno} & 0.07054 $\pm$ 0.00164 \\
 & FNO + KITINet & \textbf{0.05113 $\boldsymbol{\pm}$ 0.00093} \\
\midrule
\multirow{2}{*}{Airfoil$^{*}$}
 & OFormer~\citep{li2023oformer} & 16.39461 $\pm$ 0.12910 \\
 & KITI-OFormer & \textbf{15.49034 $\boldsymbol{\pm}$ 0.10939} \\
\bottomrule
\end{tabular}%
}
\begin{minipage}{\linewidth}
\footnotesize
$^{*}$ Airfoil problem uses the Relative Root MSE measurement.
\end{minipage}
\end{table}

\textbf{Dataset and models.} For PDE tasks, we consider benchmark equation families with varying discretizations to assess resolution generalization. Our datasets are generated following the procedure in \cref{app:pde_data}. We select the Fourier Neural Operator (FNO)~\citep{li2021fno} and Operator Transformer (OFormer)~\citep{li2023oformer} as the neural solvers. Further details are provided in  \cref{app:pde_solver}.

\textbf{Settings.}
FNOs are trained using Adam with an initial learning rate of $ 10^{-3}$, a batch size of $20$, and 1K epochs. OFormers are trained using Adam with an initial learning rate of $10^{-3}$, a batch size of $16$, and 50K epochs.   

\textbf{Results.}
\rev{\cref{Tab:pde_exp} compares vanilla and KITINet-augmented neural operators. Relative to FNO, KITI-FNO reduces the reported relative MSE by $23.5\%$ on Burgers' equation, $5.6\%$ on Navier--Stokes, and $27.5\%$ on the heat equation. Relative to OFormer, KITI-OFormer reduces the reported airfoil relative root MSE by $5.5\%$. \cref{fig:ns} provides qualitative prediction and error maps for the Navier--Stokes case.}

% \textbf{Why PDE tasks benefit more.} The PDE tasks show the largest relative gains (up to $27.5\%$), which we attribute to two factors. First, these tasks involve learning smooth solution operators where high-frequency noise in the residual stream is particularly damaging to generalization across resolutions. Second, PDE datasets are relatively small (1,000 training samples for Burgers' and NS, 2,000 for heat), making them more susceptible to overfitting. KITINet's collision-based regularization is especially effective in this limited-data regime. We note that evaluating on normalized error metrics (e.g., $L^2$ relative error normalized by the solution norm) would provide more task-agnostic comparisons; the values in~\cref{Tab:pde_exp} use the standard FNO evaluation protocol~\citep{li2021fno} and are consistent with published results.

\subsection{Controlled Studies}
\label{subsec:controlled_studies}
\subsubsection{Comparison with Alternative Stochastic Training  Strategies}\par\rev{\textbf{Setup.} We compare KITINet with three representative alternatives on the FNO backbone: Dropout~\citep{JMLR:v15:srivastava14a}, sharpness-aware minimization (SAM)~\citep{foret2021sharpness}, and stochastic depth~\citep{huang2016deep}. All three are stochastic training strategies in the evaluated setting: Dropout and stochastic depth apply random masks to activations or residual branches, whereas SAM uses stochastic mini-batch gradients to optimize against local parameter perturbations. All variants use the same datasets, FNO architecture, training budget, and evaluation protocol as the corresponding experiments in \cref{subsec:pde solving}; only the training strategy is changed.}\par

\begin{table}[t]
\centering
\caption{\rev{Comparison of FNO training strategies on the Navier--Stokes and heat equations. Values are reported as mean $\pm$ standard error.}}
\label{Tab:comparison_exp}
\fontsize{10pt}{12pt}\selectfont
\renewcommand\arraystretch{0.95}\begin{tabular*}{\linewidth}{@{\extracolsep{\fill}}lcc@{}}
\toprule
Equation & Training strategy & Relative MSE$\downarrow$ \\
\midrule
\multirow{5}{*}{NS}
 & FNO & 0.12023 $\pm$ 0.00587 \\
 & FNO + Dropout & 0.12143 $\pm$ 0.00538 \\
 & FNO + SAM & 0.13739 $\pm$ 0.00867 \\
 & FNO + Stoch. Depth & 0.14717 $\pm$ 0.00772 \\
 & KITI-FNO & \textbf{0.11346 $\boldsymbol{\pm}$ 0.00717} \\
\midrule
\multirow{5}{*}{Heat}
 & FNO & 0.07054 $\pm$ 0.00164 \\
 & FNO + Dropout & 0.07144 $\pm$ 0.00119 \\
 & FNO + SAM & 0.07444 $\pm$ 0.00124 \\
 & FNO + Stoch. Depth & 0.06655 $\pm$ 0.00078 \\
 & KITI-FNO & \textbf{0.05113 $\boldsymbol{\pm}$ 0.00093} \\
\bottomrule
\end{tabular*}
\end{table}

\rev{\textbf{Results.} \cref{Tab:comparison_exp} shows that FNO + KITINet attains the lowest mean relative MSE on both equations. On Navier--Stokes, it reduces the mean error by $5.6\%$ relative to vanilla FNO, whereas Dropout, SAM, and stochastic depth all increase it. On the heat equation, stochastic depth improves upon vanilla FNO, but KITI-FNO achieves a larger $27.5\%$ reduction and improves over stochastic depth by $23.2\%$. Across the two controlled comparisons, KITINet is the only tested strategy that consistently improves FNO and ranks first on both equations.}\par\subsubsection{Ablation of the Position Update}
\label{subsubsec:Ablation}

\begin{figure}[t]
  \centering
  \begin{subfigure}[b]{0.8\columnwidth}
    \includegraphics[width=\linewidth]{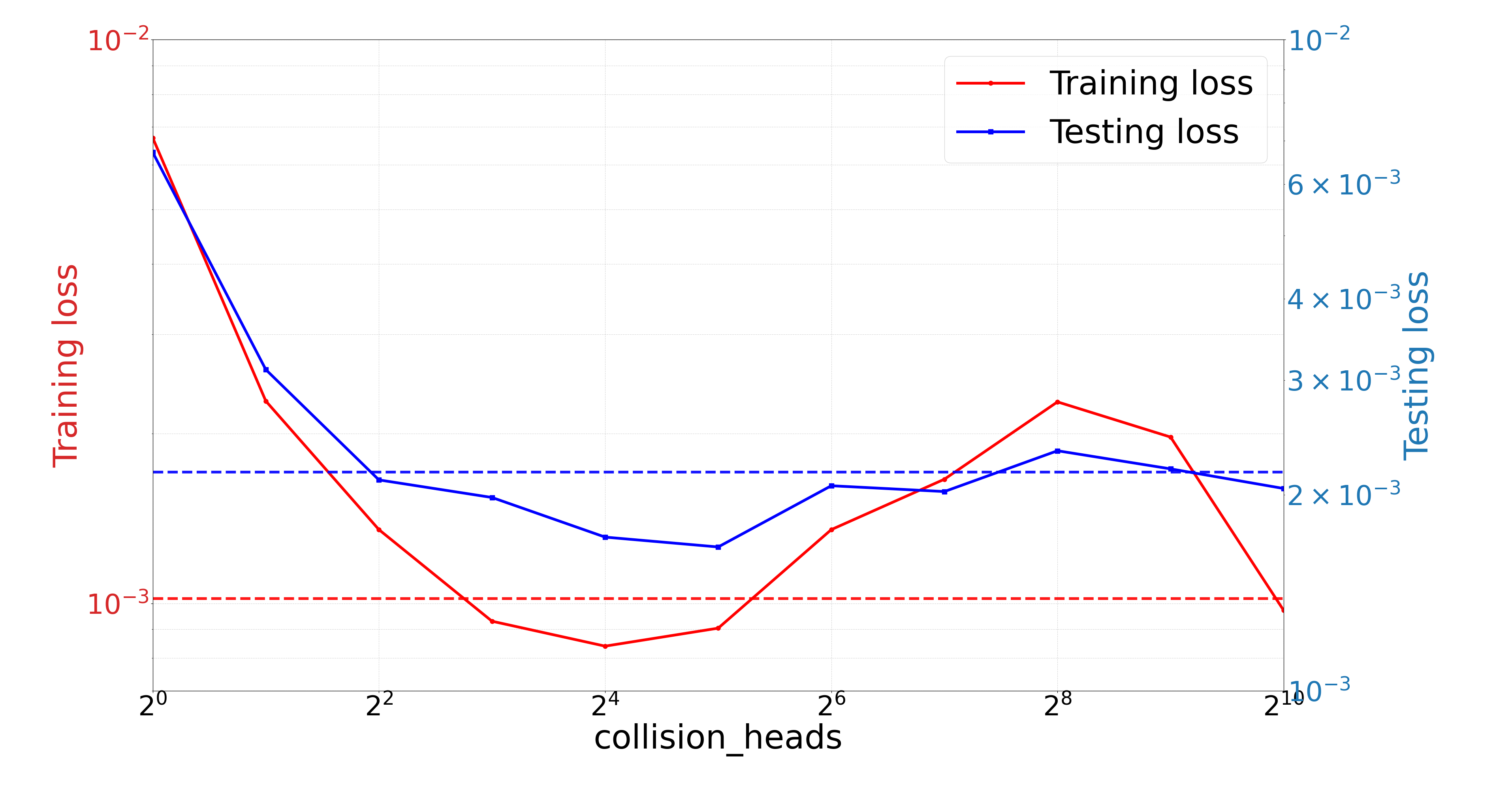}
    \caption{Burgers' Equation}
    \label{fig:burgers_loss}
  \end{subfigure}
  \hfill
  \begin{subfigure}[b]{0.8\columnwidth}
    \includegraphics[width=\linewidth]{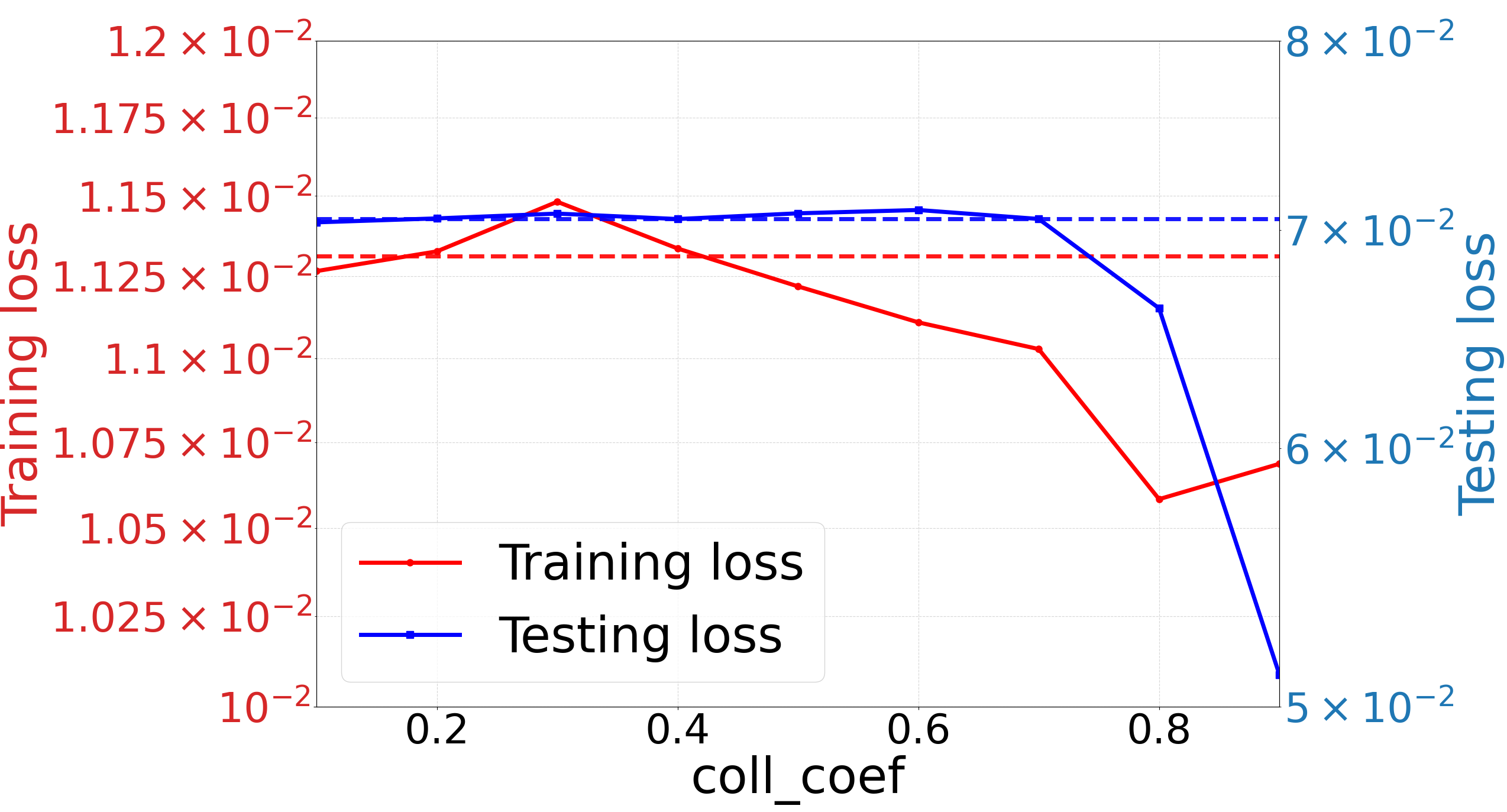}
    \caption{Heat Equation}
    \label{fig:heat_loss}
  \end{subfigure}
  \caption{KITI-FNO with different hyper-parameters $H$ and $\mathrm{coll\_coef}$ on Burgers' and Heat equations. The red and blue dashed lines show the vanilla FNO baselines. The sweep identifies task-specific operating ranges in which the hyperparameters yield substantial improvements over the vanilla FNO baselines.}
  \label{fig:hyperparameter}
\end{figure}

\rev{Classical DSMC treats the position change during a sufficiently small collision step as negligible. KITINet instead uses $\mathrm{d}t=1$ to match the scale of a residual update, so we introduce the explicit averaged position term $\boldsymbol{x}_{i}^{*}$. The averaged position term carries this residual-scale displacement into the interaction update.}
% with its original position, as described in~\cref{eq: xv_new}.

\rev{For FNO, we ablate the position update while keeping the other settings fixed. In \cref{Tab:ablation_exp}, the update variant has lower relative MSE on all three tested equations. The consistent reductions across all three equations demonstrate the utility of the position update.}

% \begin{table}[t]
% \centering
% 
% \caption{Comparing update and non-update position mechanism for KITI-FNO.}
% \label{Tab:ablation_exp}
% 
% \renewcommand\arraystretch{1.0}

% \begin{tabular}{lcc}
% \toprule
% Equation & Mechanism & MSE $\downarrow$ \\
% \midrule
% \multirow{2}{*}{Burgers'}
%  & non-update & 0.00173 \\
%  & update     & \fst{0.00166} \\
% \midrule
% \multirow{2}{*}{NS}
%  & non-update & 0.11429 \\
%  & update     & \fst{0.11346} \\
% \midrule
% \multirow{2}{*}{Heat}
%  & non-update & 0.05466 \\
%  & update     & \fst{0.05113} \\
% \bottomrule
% \end{tabular}

% 
% \end{table}
\begin{table}[t]
\centering
\caption{Update vs non-update position mechanism for KITI-FNO. \rev{Values are reported as mean $\pm$ standard error.}}
\label{Tab:ablation_exp}
\fontsize{10pt}{12pt}\selectfont
\renewcommand\arraystretch{0.95}\begin{tabular*}{0.95\linewidth}{@{\extracolsep{\fill}}lcc@{}}
\toprule
Equation & Mechanism & Relative MSE$\downarrow$ \\
\midrule
\multirow{2}{*}{Burgers'}
 & non-update & 0.00173 $\pm$ 0.00003 \\
 & update     & \textbf{0.00166 $\boldsymbol{\pm}$ 0.00005} \\
\midrule
\multirow{2}{*}{NS}
 & non-update & 0.11429 $\pm$ 0.01130 \\
 & update     & \textbf{0.11346 $\boldsymbol{\pm}$ 0.00717} \\
\midrule
\multirow{2}{*}{Heat}
 & non-update & 0.05466 $\pm$ 0.00119\\
 & update     & \textbf{0.05113 $\boldsymbol{\pm}$ 0.00093} \\
\bottomrule
\end{tabular*}
\end{table}

\subsubsection{Hyperparameter Sensitivity}
\rev{We further examine the number of collision heads $H$ and the collision coefficient $\text{coll}\_{\textrm{coef}}$. For Burgers' equation, sweeping $H$ from $1$ to $2^{10}$ identifies a best-performing range for the collision-head dimension. Sweeping $\text{coll}\_{\textrm{coef}}$ from $0.1$ to $0.9$ on the Navier--Stokes and heat equations likewise yields task-dependent optima. These sweeps identify task-dependent optima and provide practical guidance for selecting both hyperparameters.}
\subsection{Further Study on Model Behavior}
\label{sec:Analysis}
\begin{figure*}[t]
    \centering
    \includegraphics[width=0.8\linewidth]{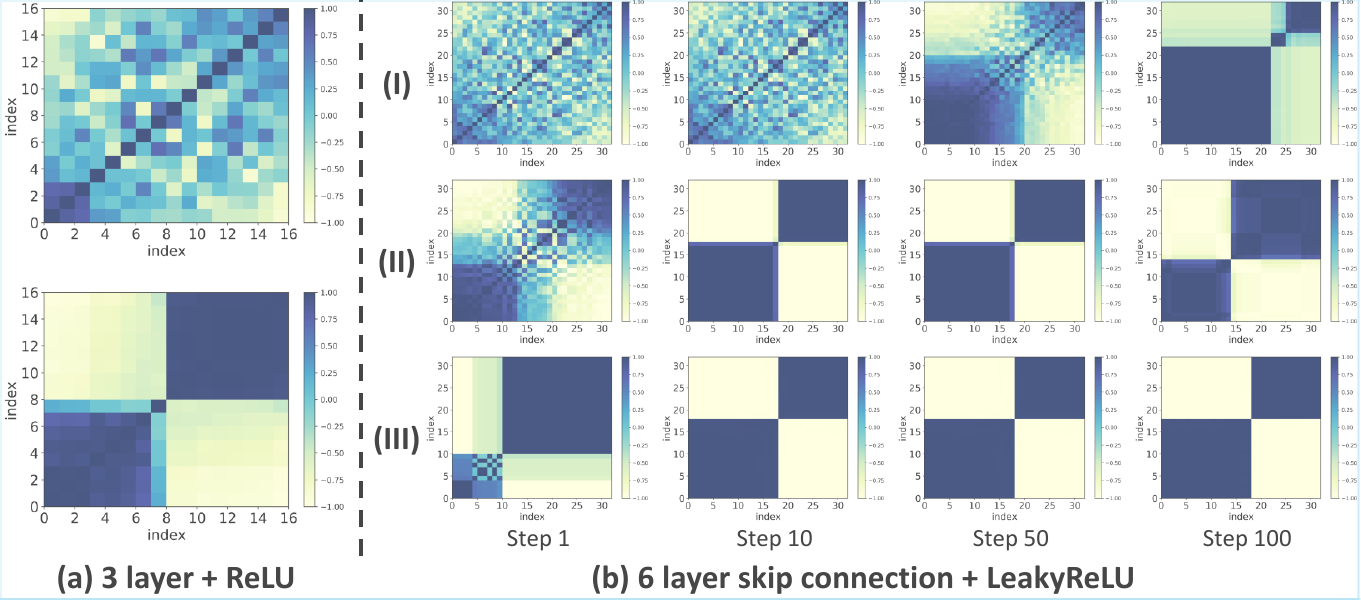}
    \caption{\rev{Parameter-condensation measurements on synthetic data. (a) Condensation patterns in three-layer FC-ReLU networks before and after replacing the final layer with KITINet. (b) Evolution in a six-layer skip-connected network with LeakyReLU when KITINet is applied to the final one or two layers. The trajectories are shown at $t \in \{1,10,50,100\}$ over 100 training epochs. In these configurations, KITINet reaches higher condensation measurements than the corresponding baseline.}}
    \label{fig:condense}
\end{figure*}
\rev{This section measures parameter condensation under KITINet in synthetic networks and in selected ResNet and FNO layers. These measurements characterize how KITINet shapes training dynamics and connect the observed condensation patterns.}

\textbf{Synthesis experiments setup.}
We consider the neural network with $d_{\text{input}}$ input dimension and $d_{\text{out}}$ output dimension, where all hidden layers have uniform width $m$. For both fully-connected and skip structures, parameters are initialized from a Gaussian distribution $N(0,\sigma)$ with $\sigma = \frac{1}{m^\gamma}$. We construct dataset from the function $\sum_{i=1}^5 3.5\sin(5x_i+1)$, where $\boldsymbol{x} = (x_1, x_2, x_3, x_4, x_5)\in[-4,2]^5$. We set $d_{\text{input}} = 5$, $d_{\text{output}}=1$, training set size $n=80$ and $\gamma=4$, following the condensation analysis framework in \citep{zhou2022towards}. To assess generalizability, we evaluate performance across multiple activation functions, including ReLU, LeakyReLU, Sigmoid, and Tanh.

\textbf{Results on a fully-connected network.}
\rev{We use a three-layer fully connected network with architecture $d_{\text{input}}$-$m$-$d_{\text{output}}$ as the baseline and replace its second linear layer with KITINet. \cref{fig:condense}(a) and \cref{evolution} show that KITINet maintains or increases the measured condensation degree across the four tested activations. For ReLU, LeakyReLU, and Tanh, the observed test-loss reductions relative to the baseline are $6.8\%$, $7.5\%$, and $4.5\%$, respectively. The increased condensation and reduced test loss connect KITINet's condensation dynamics with improved generalization in the synthetic setting.}

\textbf{Results on skip-connection neural network.}
\rev{Skip connections are central to modern deep networks~\citep{he2016deep, li2023oformer, Devlin2019BERTPO}. We use a six-layer skip-connected baseline and replace either the final layer or the final two layers with KITINet. As shown in \cref{fig:condense}(b), both variants exhibit faster condensation than the baseline, and the two-layer variant condenses faster than the one-layer variant in this experiment. The stronger effect with two insertion points indicates that condensation accumulates across KITINet layers. Additional trajectories are provided in \cref{Appendix condensation}.}

\begin{table}[t]
\centering
\caption{Condensation degree $\rho$ of the Navier--Stokes equation for vanilla FNO and KITINet.}
\label{Tab: condensation layer}
\setlength{\tabcolsep}{2.5pt}
\renewcommand\arraystretch{0.92}
\resizebox{0.95\linewidth}{!}{%
\begin{tabular}{lcccccc}
\toprule 
Layer & Conv0 & Conv1 & Conv2 & Linear0 & Linear1 & Linear2 \\
\midrule
FNO &
\textbf{0.113} & 0.089 & 0.092 & 0.188 & 0.175 & 0.182 \\
\midrule
KITI-FNO &
0.102 & \textbf{0.164} & \textbf{0.130} & \textbf{0.206} & \textbf{0.192} & \textbf{0.191} \\
\bottomrule
\end{tabular}%
}
\end{table}
\rev{\textbf{Results on Real-world Experiments.} We measure the condensation degree $\rho$ using average absolute cosine similarity~\citep{Jin2020HowDW}; the formal definition is provided in \cref{Sec: difinition of degree}. On the Navier--Stokes task, \cref{Tab: condensation layer} shows higher condensation for KITI-FNO in five of the six reported layers. On CIFAR-100, the first convolutional layer of ResNet18 increases from $0.235$ to $0.248$, whereas deeper layers change little. These measurements demonstrate that KITINet strengthens condensation across operator-learning and image-classification models.}

% \begin{table}[t]

% \caption{Condensation degree $\rho$ of the Navier--Stokes equation for vanilla FNO and KITINet.}
% 
% \centering
% \resizebox{\linewidth}{!}{
% \begin{tabular}{lcccccc}
% \toprule 
% Layer & Conv0 & Conv1 & Conv2 & Linear0 & Linear1 & Linear2 \\
% \midrule
% FNO &
% \textbf{0.113} & 0.089 & 0.092 & 0.188 & 0.175 & 0.182 \\
% \midrule
% KITI-FNO &
% 0.102 & \textbf{0.164} & \textbf{0.130} & \textbf{0.206} & \textbf{0.192} & \textbf{0.191} \\
% \bottomrule
% \end{tabular}
% }
% 
% \label{Tab: condensation layer}
% \end{table}

\subsection{\texorpdfstring{\rev{Limitations}}{Limitations}}
\rev{KITINet adds no additional trainable parameters and incurs no additional inference cost. Its all-pairs interaction has training-time complexity $O(D^2/H)$ and uses $H$ and $\mathrm{coll\_coef}$. Performance depends on task, architecture, and hyperparameter selection, motivating automated tuning. Our experiments focus on selected insertion points, and broader placement studies are a natural next step. The theoretical analysis uses a surrogate stochastic perturbation model to isolate condensation dynamics; future work will develop tighter analyses of the practical operator.}
\section{Conclusion and Future Work}

\rev{We introduced KITINet, a training-time module with no additional trainable parameters that augments residual updates with structured stochastic pairwise feature interactions inspired by DSMC. The module is removed at inference, so deployment uses the original backbone and incurs no additional inference cost. Across the evaluated language, vision, and PDE settings, KITINet demonstrates broad performance gains over the corresponding backbones. We also observe stronger parameter condensation in several settings. A simplified stochastic model provides a mechanistic account of how collision-inspired interactions promote this behavior. Future work will improve the efficiency of the pairwise interaction, develop automated hyperparameter selection, and extend the proposed mechanism across broader architectures and training regimes.}

% \section*{References}
\begin{acks}
This work was in part supported by the National Natural Science Foundation
of China (NSFC) under Grants 92370201 and 62222607.
\end{acks}
\bibliographystyle{ACM-Reference-Format}
\bibliography{kitinet}

@InProceedings{ai2winogrande,
title = {WinoGrande: An Adversarial Winograd Schema Challenge at Scale},
authors={Keisuke, Sakaguchi and Ronan, Le Bras and Chandra, Bhagavatula and Yejin, Choi
},
year={2019}
}

@inproceedings{zellers2019hellaswag,
    title={HellaSwag: Can a Machine Really Finish Your Sentence?},
    author={Zellers, Rowan and Holtzman, Ari and Bisk, Yonatan and Farhadi, Ali and Choi, Yejin},
    booktitle ={Proceedings of the 57th Annual Meeting of the Association for Computational Linguistics},
    year={2019}
}

@article{allenaiarc,
      author    = {Peter Clark  and Isaac Cowhey and Oren Etzioni and Tushar Khot and
                    Ashish Sabharwal and Carissa Schoenick and Oyvind Tafjord},
      title     = {Think you have Solved Question Answering? Try ARC, the AI2 Reasoning Challenge},
      journal   = {arXiv:1803.05457v1},
      year      = {2018},
}

@article{hendryckstest2021,
      title={Measuring Massive Multitask Language Understanding},
      author={Dan Hendrycks and Collin Burns and Steven Basart and Andy Zou and Mantas Mazeika and Dawn Song and Jacob Steinhardt},
      journal={Proceedings of the International Conference on Learning Representations (ICLR)},
      year={2021}
    }

@inproceedings{Huang2024OpenCoderTO,
  title = {OpenCoder: The Open Cookbook for Top-Tier Code Large Language Models},
  author = {Siming Huang and Tianhao Cheng and Jason Klein Liu and Jiaran Hao and Liuyihan Song and Yang Xu and J. Yang and J. H. Liu and Chenchen Zhang and Linzheng Chai and Ruifeng Yuan and Zhaoxiang Zhang and Jie Fu and Qian Liu and Ge Zhang and Zili Wang and Yuan Qi and Yinghui Xu and Wei Chu},
  year = {2024},
  url = {https://arxiv.org/pdf/2411.04905}
}

@article{zhou2025megamath,
  title     = {MegaMath: Pushing the Limits of Open Math Corpora},
  author    = {Zhou, Fan and Wang, Zengzhi and Ranjan, Nikhil and Cheng, Zhoujun and Tang, Liping and He, Guowei and Liu, Zhengzhong and Xing, Eric P.},
  journal   = {arXiv preprint arXiv:2504.02807},
  year      = {2025},
  note      = {Preprint}
}

@misc{lozhkov2024fineweb-edu,
    author       = { Lozhkov, Anton and Ben Allal, Loubna and von Werra, Leandro and Wolf, Thomas },  
    title        = { FineWeb-Edu: the Finest Collection of Educational Content }, 
    year         = 2024,  
    url          = { https://huggingface.co/datasets/HuggingFaceFW/fineweb-edu },  
    doi          = { 10.57967/hf/2497 },
    publisher    = { Hugging Face }
}

@article{lewkowycz2022solving,
  title={Solving quantitative reasoning problems with language models},
  author={Lewkowycz, Aitor and Andreassen, Anders and Dohan, David and Dyer, Ethan and Michalewski, Henryk and Ramasesh, Vinay and Slone, Ambrose and Anil, Cem and Schlag, Imanol and Gutman-Solo, Theo and others},
  journal={Advances in neural information processing systems},
  volume={35},
  pages={3843--3857},
  year={2022}
}

@inproceedings{amini-etal-2019-mathqa,
    title = "{M}ath{QA}: Towards Interpretable Math Word Problem Solving with Operation-Based Formalisms",
    author = "Amini, Aida  and
      Gabriel, Saadia  and
      Lin, Shanchuan  and
      Koncel-Kedziorski, Rik  and
      Choi, Yejin  and
      Hajishirzi, Hannaneh",
    booktitle = "Proceedings of the 2019 Conference of the North {A}merican Chapter of the Association for Computational Linguistics: Human Language Technologies, Volume 1 (Long and Short Papers)",
    month = jun,
    year = "2019",
    address = "Minneapolis, Minnesota",
    publisher = "Association for Computational Linguistics",
    url = "https://aclanthology.org/N19-1245",
    doi = "10.18653/v1/N19-1245",
    pages = "2357--2367",
}

@article{hendrycksmath2021,
  title={Measuring Mathematical Problem Solving With the MATH Dataset},
  author={Dan Hendrycks and Collin Burns and Saurav Kadavath and Akul Arora and Steven Basart and Eric Tang and Dawn Song and Jacob Steinhardt},
  journal={NeurIPS},
  year={2021}
}

@article{cobbe2021gsm8k,
  title={Training Verifiers to Solve Math Word Problems},
  author={Cobbe, Karl and Kosaraju, Vineet and Bavarian, Mohammad and Chen, Mark and Jun, Heewoo and Kaiser, Lukasz and Plappert, Matthias and Tworek, Jerry and Hilton, Jacob and Nakano, Reiichiro and Hesse, Christopher and Schulman, John},
  journal={arXiv preprint arXiv:2110.14168},
  year={2021}
}

@misc{paster2023openwebmath,
      title={OpenWebMath: An Open Dataset of High-Quality Mathematical Web Text},
      author={Keiran Paster and Marco Dos Santos and Zhangir Azerbayev and Jimmy Ba},
      year={2023},
      eprint={2310.06786},
      archivePrefix={arXiv},
      primaryClass={cs.AI}
}

@article{JMLR:v15:srivastava14a,
  author  = {Nitish Srivastava and Geoffrey Hinton and Alex Krizhevsky and Ilya Sutskever and Ruslan Salakhutdinov},
  title   = {Dropout: A Simple Way to Prevent Neural Networks from Overfitting},
  journal = {Journal of Machine Learning Research},
  year    = {2014},
  volume  = {15},
  number  = {56},
  pages   = {1929--1958},
  url     = {http://jmlr.org/papers/v15/srivastava14a.html}
}

@article{Maennel2018GradientDQ,
  title={Gradient Descent Quantizes ReLU Network Features},
  author={Hartmut Maennel and Olivier Bousquet and Sylvain Gelly},
  journal={ArXiv},
  year={2018},
  volume={abs/1803.08367},
  url={https://api.semanticscholar.org/CorpusID:4050418}
}

@article{Lyu2021GradientDO,
  title={Gradient Descent on Two-layer Nets: Margin Maximization and Simplicity Bias},
  author={Kaifeng Lyu and Zhiyuan Li and Runzhe Wang and Sanjeev Arora},
  journal={ArXiv},
  year={2021},
  volume={abs/2110.13905},
  url={https://api.semanticscholar.org/CorpusID:239885353}
}

@article{Min2023EarlyNA,
  title={Early Neuron Alignment in Two-layer ReLU Networks with Small Initialization},
  author={Hancheng Min and Ren{\'e} Vidal and Enrique Mallada},
  journal={ArXiv},
  year={2023},
  volume={abs/2307.12851},
  url={https://api.semanticscholar.org/CorpusID:260125817}
}

@article{Ma2024ConvergenceOG,
  title={Convergence of Gradient Descent with Small Initialization for Unregularized Matrix Completion},
  author={Jianhao Ma and Salar Fattahi},
  journal={ArXiv},
  year={2024},
  volume={abs/2402.06756},
  url={https://api.semanticscholar.org/CorpusID:267627789}
}

@article{Jin2020HowDW,
  title={How does Weight Correlation Affect the Generalisation Ability of Deep Neural Networks},
  author={Gao Jin and Xinping Yi and Liang Zhang and Lijun Zhang and Sven Schewe and Xiaowei Huang},
  journal={ArXiv},
  year={2020},
  volume={abs/2010.05983},
  url={https://api.semanticscholar.org/CorpusID:222310564}
}

@inproceedings{he2016deep,
  title={Deep residual learning for image recognition},
  author={He, Kaiming and Zhang, Xiangyu and Ren, Shaoqing and Sun, Jian},
  booktitle={Computer Vision and Pattern Recognition},
  pages={770--778},
  year={2016}
}

@article{vaswani2017attention,
  title={Attention is all you need},
  author={Vaswani, Ashish and Shazeer, Noam and Parmar, Niki and Uszkoreit, Jakob and Jones, Llion and Gomez, Aidan N and Kaiser, {\L}ukasz and Polosukhin, Illia},
  journal={Advances in neural information processing systems},
  volume={30},
  year={2017}
}

@article{xu2025overview,
  title={An overview of condensation phenomenon in deep learning},
  author={Xu, Zhi-Qin John and Zhang, Yaoyu and Zhou, Zhangchen},
  journal={arXiv preprint arXiv:2504.09484},
  year={2025}
}

@article{abe1993generalized,
  title={Generalized scheme of the no-time-counter scheme for the DSMC in rarefied gas flow analysis},
  author={Abe, Takashi},
  journal={Computers \& fluids},
  volume={22},
  number={2-3},
  pages={253--257},
  year={1993},
  publisher={Elsevier}
}

@article{zhou2022towards,
  title={Towards understanding the condensation of neural networks at initial training},
  author={Zhou, Hanxu and Qixuan, Zhou and Luo, Tao and Zhang, Yaoyu and Xu, Zhi-Qin},
  journal={Advances in Neural Information Processing Systems},
  volume={35},
  pages={2184--2196},
  year={2022}
}

@article{chen2024efficient,
  title={Efficient and Flexible Method for Reducing Moderate-Size Deep Neural Networks with Condensation},
  author={Chen, Tianyi and Xu, Zhi-Qin John},
  journal={Entropy},
  volume={26},
  number={7},
  pages={567},
  year={2024},
  publisher={MDPI}
}

@article{zhang2021embedding,
  title={Embedding principle of loss landscape of deep neural networks},
  author={Zhang, Yaoyu and Zhang, Zhongwang and Luo, Tao and Xu, Zhiqin J},
  journal={Advances in Neural Information Processing Systems},
  volume={34},
  pages={14848--14859},
  year={2021}
}

@article{hollingsworth2018molecular,
  title={Molecular dynamics simulation for all},
  author={Hollingsworth, Scott A and Dror, Ron O},
  journal={Neuron},
  volume={99},
  number={6},
  pages={1129--1143},
  year={2018},
  publisher={Elsevier}
}

@article{boltzmann2015relationship,
  title={On the relationship between the second fundamental theorem of the mechanical theory of heat and probability calculations regarding the conditions for thermal equilibrium},
  author={Boltzmann, Ludwig},
  journal={Entropy},
  volume={17},
  number={4},
  pages={1971--2009},
  year={2015}
}

@article{bird1963approach,
  title={Approach to translational equilibrium in a rigid sphere gas},
  author={Bird, GA},
  journal={Phys. fluids},
  volume={6},
  pages={1518--1519},
  year={1963}
}

@article{eliasof2021pde,
  title={Pde-gcn: Novel architectures for graph neural networks motivated by partial differential equations},
  author={Eliasof, Moshe and Haber, Eldad and Treister, Eran},
  journal={Advances in neural information processing systems},
  volume={34},
  pages={3836--3849},
  year={2021}
}

@article{kidger2020neural,
  title={Neural controlled differential equations for irregular time series},
  author={Kidger, Patrick and Morrill, James and Foster, James and Lyons, Terry},
  journal={Advances in neural information processing systems},
  volume={33},
  pages={6696--6707},
  year={2020}
}

@article{rubanova2019latent,
  title={Latent ordinary differential equations for irregularly-sampled time series},
  author={Rubanova, Yulia and Chen, Ricky TQ and Duvenaud, David K},
  journal={Advances in neural information processing systems},
  volume={32},
  year={2019}
}

@article{schuetz2022combinatorial,
  title={Combinatorial optimization with physics-inspired graph neural networks},
  author={Schuetz, Martin JA and Brubaker, J Kyle and Katzgraber, Helmut G},
  journal={Nature Machine Intelligence},
  volume={4},
  number={4},
  pages={367--377},
  year={2022},
}

@article{long2019pde,
  title={PDE-Net 2.0: Learning PDEs from data with a numeric-symbolic hybrid deep network},
  author={Long, Zichao and Lu, Yiping and Dong, Bin},
  journal={Journal of Computational Physics},
  volume={399},
  pages={108925},
  year={2019},
}

@inproceedings{norcliffe2021neural,
  title={Neural ODE Processes},
  author={Norcliffe, A and Bodnar, C and Day, B and Moss, J and Lio, P and others},
  booktitle={International Conference on Learning Representations},
  year={2021},
}

@article{norcliffe2020second,
  title={On second order behaviour in augmented neural odes},
  author={Norcliffe, Alexander and Bodnar, Cristian and Day, Ben and Simidjievski, Nikola and Li{\`o}, Pietro},
  journal={Advances in neural information processing systems},
  volume={33},
  pages={5911--5921},
  year={2020}
}

@inproceedings{long2018pde,
  title={Pde-net: Learning pdes from data},
  author={Long, Zichao and Lu, Yiping and Ma, Xianzhong and Dong, Bin},
  booktitle={International conference on machine learning},
  pages={3208--3216},
  year={2018},
  organization={PMLR}
}

@article{wang2025convection,
  title={Convection-Diffusion Equation: a Theoretically Certified Framework for Neural Networks},
  author={Wang, Tangjun and Bao, Chenglong and Shi, Zuoqiang},
  journal={IEEE Transactions on Pattern Analysis and Machine Intelligence},
  year={2025},
}

@article{greydanus2019hamiltonian,
  title={Hamiltonian neural networks},
  author={Greydanus, Samuel and Dzamba, Misko and Yosinski, Jason},
  journal={Advances in neural information processing systems},
  volume={32},
  year={2019}
}

@article{toth2019hamiltonian,
  title={Hamiltonian generative networks},
  author={Toth, Peter and Rezende, Danilo Jimenez and Jaegle, Andrew and Racani{\`e}re, S{\'e}bastien and Botev, Aleksandar and Higgins, Irina},
  journal={arXiv preprint arXiv:1909.13789},
  year={2019}
}

@article{bilovs2021neural,
  title={Neural flows: Efficient alternative to neural ODEs},
  author={Bilo{\v{s}}, Marin and Sommer, Johanna and Rangapuram, Syama Sundar and Januschowski, Tim and G{\"u}nnemann, Stephan},
  journal={Advances in neural information processing systems},
  volume={34},
  pages={21325--21337},
  year={2021}
}

@article{chen2018neural,
  title={Neural ordinary differential equations},
  author={Chen, Ricky TQ and Rubanova, Yulia and Bettencourt, Jesse and Duvenaud, David K},
  journal={Advances in neural information processing systems},
  volume={31},
  year={2018}
}

@misc{li2021fno,
      title={Fourier Neural Operator for Parametric Partial Differential Equations}, 
      author={Zongyi Li and Nikola Kovachki and Kamyar Azizzadenesheli and Burigede Liu and Kaushik Bhattacharya and Andrew Stuart and Anima Anandkumar},
      year={2021},
      eprint={2010.08895},
      archivePrefix={arXiv},
      primaryClass={cs.LG},
      url={https://arxiv.org/abs/2010.08895}, 
}

@misc{li2023oformer,
      title={Transformer for Partial Differential Equations' Operator Learning}, 
      author={Zijie Li and Kazem Meidani and Amir Barati Farimani},
      year={2023},
      eprint={2205.13671},
      archivePrefix={arXiv},
      primaryClass={cs.LG},
      url={https://arxiv.org/abs/2205.13671}, 
}

@article{su2,
	title = {{SU}2: {An} {Open}-{Source} {Suite} for {Multiphysics} {Simulation} and {Design}},
	volume = {54},
	url = {https://doi.org/10.2514/1.J053813},
	doi = {10.2514/1.J053813},
	number = {3},
	journal = {AIAA Journal},
	author = {Economon, Thomas D. and Palacios, Francisco and Copeland, Sean R. and Lukaczyk, Trent W. and Alonso, Juan J.},
	month = dec,
	year = {2015},
	pages = {828--846}
}

@misc{Mesh-Based,
      title={Learning Mesh-Based Simulation with Graph Networks}, 
      author={Tobias Pfaff and Meire Fortunato and Alvaro Sanchez-Gonzalez and Peter W. Battaglia},
      year={2021},
      eprint={2010.03409},
      archivePrefix={arXiv},
      primaryClass={cs.LG},
      url={https://arxiv.org/abs/2010.03409}, 
}

@article{liu2024dendritic,
  title={Dendritic Integration Inspired Artificial Neural Networks Capture Data Correlation},
  author={Liu, Chongming and Ma, Jingyang and Li, Songting and Zhou, Douglas Dongzhuo},
  journal={Advances in Neural Information Processing Systems},
  volume={37},
  pages={79325--79349},
  year={2024}
}

@inproceedings{bowman-etal-2015-large,
    title = "A large annotated corpus for learning natural language inference",
    author = "Bowman, Samuel R.  and
      Angeli, Gabor  and
      Potts, Christopher  and
      Manning, Christopher D.",
    editor = "M{\`a}rquez, Llu{\'\i}s  and
      Callison-Burch, Chris  and
      Su, Jian",
    booktitle = "Proceedings of the 2015 Conference on Empirical Methods in Natural Language Processing",
    month = sep,
    year = "2015",
    address = "Lisbon, Portugal",
    publisher = "Association for Computational Linguistics",
    url = "https://aclanthology.org/D15-1075",
    doi = "10.18653/v1/D15-1075",
    pages = "632--642",
}

@InProceedings{maas-EtAl:2011:ACL-HLT2011,
  author    = {Maas, Andrew L.  and  Daly, Raymond E.  and  Pham, Peter T.  and  Huang, Dan  and  Ng, Andrew Y.  and  Potts, Christopher},
  title     = {Learning Word Vectors for Sentiment Analysis},
  booktitle = {Proceedings of the 49th Annual Meeting of the Association for Computational Linguistics: Human Language Technologies},
  month     = {June},
  year      = {2011},
  address   = {Portland, Oregon, USA},
  publisher = {Association for Computational Linguistics},
  pages     = {142--150},
  url       = {http://www.aclweb.org/anthology/P11-1015}
}

@inproceedings{Devlin2019BERTPO,
  title={BERT: Pre-training of Deep Bidirectional Transformers for Language Understanding},
  author={Jacob Devlin and Ming-Wei Chang and Kenton Lee and Kristina Toutanova},
  booktitle={North American Chapter of the Association for Computational Linguistics},
  year={2019},
  url={https://api.semanticscholar.org/CorpusID:52967399}
}

@article{gpt2,
  title     = {Language Models Are Unsupervised Multitask Learners},
  author    = {Radford, Alec and Wu, Jeff and Child, Rewon and Luan, David and Amodei, Dario and Sutskever, Ilya},
  year      = {2019},
  journal   = {OpenAI Blog},
  volume    = {1},
  number    = {8},
  pages     = {9},
  url       = {https://openai.com/blog/better-language-models/},
}

@article{Karabulut_2007,
   title={Direct simulation for a homogeneous gas},
   volume={75},
   ISSN={1943-2909},
   url={http://dx.doi.org/10.1119/1.2366735},
   DOI={10.1119/1.2366735},
   number={1},
   journal={American Journal of Physics},
   publisher={American Association of Physics Teachers (AAPT)},
   author={Karabulut, Hasan},
   year={2007},
   month=jan, pages={62–66} }

@article{doi:10.1143/JPSJ.49.2042,
author = {Nanbu ,Kenichi},
title = {Direct Simulation Scheme Derived from the Boltzmann Equation. I. Monocomponent Gases},
journal = {Journal of the Physical Society of Japan},
volume = {49},
number = {5},
pages = {2042-2049},
year = {1980},
doi = {10.1143/JPSJ.49.2042},
URL = {https://doi.org/10.1143/JPSJ.49.2042}}

@article{Babovsky1986SimulationScheme,
  author    = {Hans Babovsky},
  title     = {On a simulation scheme for the Boltzmann equation},
  journal   = {Mathematical Methods in the Applied Sciences},
  volume    = {8},
  number    = {3},
  pages     = {223--233},
  year      = {1986},
  doi       = {10.1002/mma.1670080114},
}

@article{Garcia2025_DSMC,
  title        = {DSMC: A Statistical Mechanics Perspective},
  author       = {Alejandro L. Garcia},
  journal      = {arXiv preprint arXiv:2501.07785},
  year         = {2025},
  url          = {https://arxiv.org/abs/2501.07785},
}

@article{Wainwright1964,
  author  = {T. Wainwright},
  title   = {Calculation of hard-sphere viscosity by means of correlation functions},
  journal = {The Journal of Chemical Physics},
  year    = {1964},
  volume  = {40},
  pages   = {2932--2937}
}

@article{10.1063/1.1740082,
    author = {Green, Melville S.},
    title = {Markoff Random Processes and the Statistical Mechanics of Time‐Dependent Phenomena. II. Irreversible Processes in Fluids},
    journal = {The Journal of Chemical Physics},
    volume = {22},
    number = {3},
    pages = {398-413},
    year = {1954},
    month = {03},
    issn = {0021-9606},
    doi = {10.1063/1.1740082},
    url = {https://doi.org/10.1063/1.1740082},
    
}

@article{doi:10.1143/JPSJ.12.570,
author = {Kubo ,Ryogo},
title = {Statistical-Mechanical Theory of Irreversible Processes. I. General Theory and Simple Applications to Magnetic and Conduction Problems},
journal = {Journal of the Physical Society of Japan},
volume = {12},
number = {6},
pages = {570-586},
year = {1957},
doi = {10.1143/JPSJ.12.570},
URL = {https://doi.org/10.1143/JPSJ.12.570},
eprint = {https://doi.org/10.1143/JPSJ.12.570}
}

@misc{xie2026mhc,
      title={mHC: Manifold-Constrained Hyper-Connections}, 
      author={Zhenda Xie and Yixuan Wei and Huanqi Cao and Chenggang Zhao and Chengqi Deng and Jiashi Li and Damai Dai and Huazuo Gao and Jiang Chang and Kuai Yu and Liang Zhao and Shangyan Zhou and Zhean Xu and Zhengyan Zhang and Wangding Zeng and Shengding Hu and Yuqing Wang and Jingyang Yuan and Lean Wang and Wenfeng Liang},
      year={2026},
      eprint={2512.24880},
      archivePrefix={arXiv},
      primaryClass={cs.CL},
      url={https://arxiv.org/abs/2512.24880}, 
}

@misc{qwen2026qwen35,
  author = {{Qwen Team}},
  title = {{Qwen3.5}: Towards Native Multimodal Agents},
  month = feb,
  year = {2026},
  url = {https://qwen.ai/blog?id=qwen3.5}
}

@misc{zhang2024chemllm,
  author = {Di Zhang and Wei Liu and Qian Tan and Jingdan Chen and Hang Yan and Yuliang Yan and Jiatong Li and Weiran Huang and Xiangyu Yue and Wanli Ouyang and Dongzhan Zhou and Shufei Zhang and Mao Su and Han-Sen Zhong and Yuqiang Li},
  title = {{ChemLLM}: A Chemical Large Language Model},
  year = {2024},
  eprint = {2402.06852},
  archivePrefix = {arXiv},
  primaryClass = {cs.AI},
  doi = {10.48550/arXiv.2402.06852}
}

@misc{mirza2025chempile,
  author = {Adrian Mirza and Nawaf Alampara and Marti{\~n}o R{\'i}os-Garc{\'i}a and Mohamed Abdelalim and Jack Butler and Bethany Connolly and Tunca Dogan and Marianna Nezhurina and B{\"u}nyamin {\c{S}}en and Santosh Tirunagari and Mark Worrall and Adamo Young and Philippe Schwaller and Michael Pieler and Kevin Maik Jablonka},
  title = {{ChemPile}: A {250GB} Diverse and Curated Dataset for Chemical Foundation Models},
  year = {2025},
  eprint = {2505.12534},
  archivePrefix = {arXiv},
  primaryClass = {cs.LG},
  doi = {10.48550/arXiv.2505.12534}
}

@misc{hu2022lora,
  author = {Hu, Edward J. and Shen, Yelong and Wallis, Phillip and Allen-Zhu, Zeyuan and Li, Yuanzhi and Wang, Shean and Wang, Lu and Chen, Weizhu},
  title = {{LoRA}: Low-Rank Adaptation of Large Language Models},
  booktitle = {International Conference on Learning Representations},
  year = {2022},
  url = {https://openreview.net/forum?id=nZeVKeeFYf9}
}

@misc{loshchilov2019decoupled,
  author = {Loshchilov, Ilya and Hutter, Frank},
  title = {Decoupled Weight Decay Regularization},
  booktitle = {International Conference on Learning Representations},
  year = {2019},
  url = {https://openreview.net/forum?id=Bkg6RiCqY7}
}

@misc{foret2021sharpness,
  author = {Foret, Pierre and Kleiner, Ariel and Mobahi, Hossein and Neyshabur, Behnam},
  title = {Sharpness-Aware Minimization for Efficiently Improving Generalization},
  booktitle = {International Conference on Learning Representations},
  year = {2021},
  url = {https://openreview.net/forum?id=6Tm1mposlrM}
}

@misc{huang2016deep,
  author = {Huang, Gao and Sun, Yu and Liu, Zhuang and Sedra, Daniel and Weinberger, Kilian Q.},
  title = {Deep Networks with Stochastic Depth},
  booktitle = {European Conference on Computer Vision},
  year = {2016},
  pages = {646--661},
  doi = {10.1007/978-3-319-46493-0_39}
}

%%%%%%%%%%%%%%%%%%%%%%%%%%%%%%%%%%%%%%%%%%%%%%%%%%%%%%%%%%%%

\clearpage
\appendix
\crefalias{section}{appendix}
\crefalias{subsection}{appendix}

\section{Molecule Dynamics and Neural ODE}

Molecular dynamics (MD)~\citep{hollingsworth2018molecular} is a numerical method that simulates the motion of molecular or atomic systems at a microscopic level. Considering $N$ particles in one-dimensional space ${1,..,N}$, where the mass of the $i$-th particle is $m_i$, and its position is $x_i$. The potential energy is $V(x_1, ..., x_N)$. According to Newton's second law, we can write a second-order ordinary differential equation (ODE)
\begin{equation}
\label{eq:newton2law}
    m_i\frac{d^2x_i}{\mathrm{d}t^2} = F_i = -\nabla_{x_i} V
\end{equation}
where $F_i$ represents the net external force acting on particle $i$, including weak interaction forces (van der Waals forces), electromagnetic forces (Coulomb forces, chemical bonds), etc. Specifically, the expression for Neural ODE (NODE)~\citep{chen2018neural} involves a first-order ODE:
\begin{equation}
\frac{d\boldsymbol{x}}{\mathrm{d}t} = v(\boldsymbol{x}(t), t, \theta)
\end{equation}
where $\boldsymbol{x}\in \mathbb{R}^D$ represents the hidden layer output; $t\in \mathbb{R}_{+}$ represents the network depth, while $v$ represents the neural network with parameters $\theta$. This can be analogized to $N$ particles moving with velocity $v$. For NODE, its corresponding Newtonian equation can be written as:
\begin{equation}
\frac{d^2 \boldsymbol{x}}{\mathrm{d}t^2} = \frac{D v}{Dt} =  (\nabla_{\boldsymbol{x}} v)v + \frac{\partial v}{\partial t} :=F/m
\end{equation}
It can be regarded as a neural molecular dynamics system, where the hidden layer output $\boldsymbol{x}$ is the particle position, and velocity $v$ is the derivative of the position parameterized by learnable parameters.

The NODE is a continuous model, implemented by discrete numerical methods, e.g., the Euler method and the Runge-Kutta method. The NODE can be trained by backpropagation through the adjoint method. It can be used to solve the problem of vanishing gradient and exploding gradient, and has wide applications in the field of machine learning, e.g., image generation, time series prediction, etc.

\section{Direct Simulation Monte Carlo (DSMC)}
\label{app:DSMC}
In DSMC, we consider the hard sphere model, where the collision probability is proportional to the relative velocity of the particle pairs:
\begin{equation}
    P_{\mathrm {coll} }[i,j]={{|\boldsymbol {v} _{i}-\boldsymbol {v} _{j}|} \over {\sum _{m=1}^{N_{\mathrm {c} }}\sum _{n=1}^{m-1}|\boldsymbol {v} _{m}-\boldsymbol {v} _{n}|}},
\end{equation}
where $N_{\mathrm {c} }$ is the number of particles in the cell; the velocity $\boldsymbol {v}$ is proportional to the momentum $\boldsymbol {p}$ if assuming the particle mass is constant. The denominator is expensive to compute, so the DSMC method uses a rejection sampling method to approximate the collision probability:
\begin{enumerate}
    \item Estimate the number of candidate collision pairs $M_{\mathrm {cand} }$ by the no-time-counter method~\citep{abe1993generalized}: 
    \begin{equation}
        M_{\mathrm {cand} }={{N_{\mathrm {c} }(N_{\mathrm {c} }-1)F_{N}\pi d^{2}v_{\mathrm {r} }^{\max }\tau } \over {2V_{\mathrm {c} }}},
    \end{equation}
    where $d$ is the particle diameter, the $v_{\mathrm {r} }^{\max }$ is the estimated maximum relative velocity, the $\tau$ is the time step, the $V_{\mathrm {c} }$ is the cell volume.
    \item Randomly select $M_{\mathrm {cand} }$ pairs of particles. For each pair $i,j$, generate a random number $\Re_1$ from the uniform distribution $U(0,1)$, and accept the collision if 
    \begin{equation}
    \begin{aligned}
    |\boldsymbol {v} _{i}-\boldsymbol {v} _{j}|/v_{\mathrm {r} }^{\max } > \Re_1.
    \end{aligned}
    \label{eq: DSMC_accept}
    \end{equation}
    \item If the collision is accepted, update the velocity of the particles according to the collision model, with position unchanged.
    \item Repeat the above steps for all cells, then proceed to the next time step.
\end{enumerate}

The hard sphere model is a hard-body collision. The particles conserve pairwise momentum and energy and scatter off in a random direction. Set the post-collision relative velocity in a polar coordinate system:
\begin{equation}
\begin{aligned}
\boldsymbol {v}_{\mathrm {r} }^{*}=v_{\mathrm {r} }[(\sin \theta \cos \phi ){\hat {\boldsymbol {x} }}+(\sin \theta \sin \phi ){\hat {\boldsymbol {y} }}+\cos \theta \,{\hat {\boldsymbol {z} }}],
\end{aligned}
\label{eq: DSMC_vr}
\end{equation}
and the angles are set as $\phi =2\pi \Re _{2}$ and $\theta =\cos ^{-1}(2\Re _{3}-1)$, where $\Re _{2}$ and $\Re _{3}$ are random numbers from the uniform distribution $U(0,1)$. Denote the center of mass velocity as $\boldsymbol {v} _{\mathrm {cm} }=(\boldsymbol {v} _{i}+\boldsymbol {v} _{j})/2$, then the post-collision velocity can be calculated as:
\begin{equation}
\begin{aligned}
\boldsymbol {v} _{i}'=\boldsymbol {v} _{\mathrm {cm} }+\boldsymbol {v} _{\mathrm {r} }^{*}/2,
\boldsymbol {v} _{j}'=\boldsymbol {v} _{\mathrm {cm} }-\boldsymbol {v} _{\mathrm {r} }^{*}/2,
\end{aligned}
\label{eq:DSMC_deltav}
\end{equation}

\section{Algorithms}

Besides the KITINet architecture we have introduced, we also experimented with an alternative architecture of KITINet, a-edition KITINet. In physics, acceleration can exhibit abrupt changes due to external forces, whereas velocity should vary continuously. Therefore, a-edition KITINet considers residual connections as position $\boldsymbol{x}$ and residuals as acceleration $\boldsymbol{a}$, requiring velocity $\boldsymbol{v}$ from the previous a-edition KITINet, and outputs $\boldsymbol{x}'$ and $\boldsymbol{v}'$. For the first layer of a-edition KITINet, $\boldsymbol{v}$ would be a random variable drawn from a Gaussian distribution, which satisfies the thermodynamic distribution. During the a-edition KITINet, the initial velocity for collision simulation would be $\boldsymbol{v} + \mathrm{d}t*\boldsymbol{a}$, and the velocity after collision $\boldsymbol{v}'$ is recorded for the next a-edition KITINet.  

From a physics perspective, the a-edition KITINet more faithfully satisfies Newton's second law with \cref{eq:newton2law} and \textbf{BTE} with \cref{bte}; from a neural-network perspective, the variable $\boldsymbol{v}$ within the network functions analogously to an RNN's hidden state, storing and propagating information. However, in experiments, the a-edition KITINet failed to deliver satisfactory results.

\section{PDE-solver Architecture}
\label{app:pde_solver}
\begin{figure}[t]
    \centering
    \includegraphics[width=0.8\linewidth]{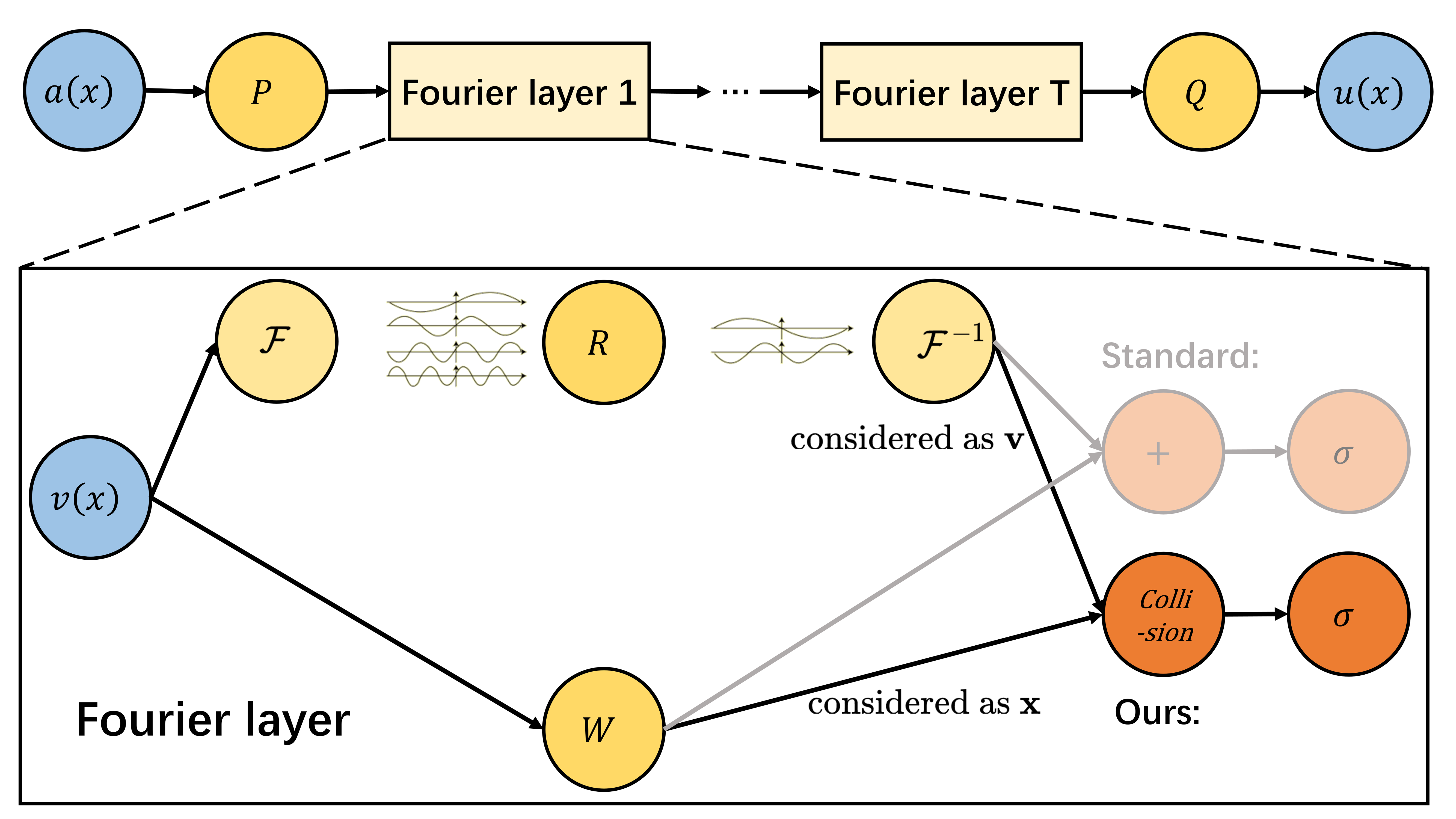}
    \caption{The full architecture of Fourier Neural Operators (FNO) with and without KITINet applied.}
    \label{fig:fno}
\end{figure}

\subsection{Fourier Neural Operator (FNO)}

FNO~\citep{li2021fno} is a neural operator that implements a resolution-invariant global convolution by FFT'ing input features, applying a learnable linear transform to a truncated set of frequency modes, and then inverse-FFT'ing back to the spatial domain. It efficiently captures long-range dependencies and generalizes across discretizations. As Fig.~\ref{fig:fno} in the appendix shows, for each Fourier layer with KITINet applied, the outputs of the Fourier convolution are considered as $\boldsymbol{v}$, while the outputs of the linear transformation are considered as $\boldsymbol{x}$.

\subsection{Operator Transformer (OFormer)}

OFormer~\citep{li2023oformer} embeds Fourier neural operator blocks into a Transformer-style sequence model, applying FFTs to input tokens, learnable complex-valued multipliers on truncated frequency modes, and inverse FFTs back to space, while its attention mechanism enables these spectral operations to be conditioned on arbitrary, irregular input locations, making it directly applicable to non-uniform and unstructured grids. In the Transformer architecture with KITINet applied, the outputs of the self-attention and MLP layers are considered as $\boldsymbol{v}$, while the residual connections are considered as $\boldsymbol{x}$.

\section{PDE Datasets Generation}
\label{app:pde_data}

\subsection{BURGERS' EQUATION}

The one-dimensional Burgers' equation is a nonlinear PDE commonly used to describe viscous fluid flow in a single spatial dimension. It takes the form (We use datasets from \citep{li2021fno}):
$$\partial_{t} u(x,t)+\partial_{x}(u^{2}(x,t)/2)=\nu\partial_{xx}u(x,t), \quad x\in(0,1),t\in(0,1],$$
$$u(x,0)=u_{0}(x),\quad x\in(0,1).$$
The initial condition $u_{0}(x)$ is generated according to $u_0 \backsim \mu$ where $\mu = \mathcal{N}(0, 625(-\Delta+25I)^{-2})$ with periodic boundary conditions, and the viscosity is set to $\nu = 0.1$. Fourier Neural Operators are chosen for solving this equation, learning the operator mapping the initial condition to the solution at time one, $G^{\dagger}: L_{per}^{2}((0,1);\mathbb{R}) \to H_{per}^{r}((0,1);\mathbb{R})$ defined by $u_{0} \mapsto u(\cdot,1)$ for any $r > 0$.

\subsection{NAVIER-STOKES (NS) EQUATION}

\begin{figure*}[!htb]
    \centering
    \includegraphics[width=0.99\linewidth]{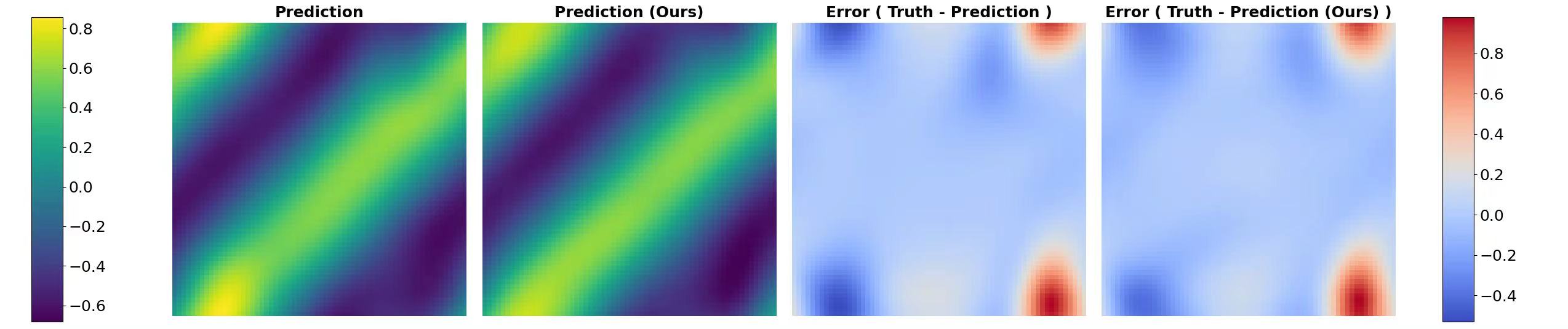}
    \caption{FNOs' performance on the NS equation, both vanilla and with KITINet applied. Left two: FNOs' predictions at the final time step; Right two: their corresponding absolute error maps.}
    \label{fig:ns}
\end{figure*}
The two-dimensional NS equation for a viscous, incompressible fluid in vorticity form on the unit torus takes the form:
$$\partial_{t}\omega(x,t)+u(x,t)\cdot\nabla\omega(x,t)=\nu\Delta\omega(x,t)+f(x),\quad x\in (0,1)^{2},t\in(0,T],$$
$$\nabla \cdot u(x,t)=0,\quad x\in (0,1)^{2},t\in(0,T],$$
$$\omega(x,0)=\omega_{0}(x), \quad x\in (0,1)^{2}.$$
The initial condition $\omega_{0}(x)$ is generated according to $\omega_0 \backsim \mu$ where $\mu = \mathcal{N}(0, 7^{3/2}(-\Delta+49I)^{-2.5})$ , with periodic boundary conditions, the force $f(x)=0.1(\sin(2\pi(x_1 + x_2)) + \cos(2\pi(x_1 + x_2)))$ and the viscosity is set to $\nu = 1$e$-3$. FNOs are chosen for this equation, learning the operator mapping the vorticity up to time 10 to the solution up to $T > 10$, $G^{\dagger}: C([0,10];H_{per}^{r}((0,1);\mathbb{R})) \to C([10,T];H_{per}^{r}((0,1));\mathbb{R})$ defined by $\omega|_{(0,1)^{2} \times (0,10]} \mapsto \omega|_{(0,1)^{2} \times (10,T]}$ for any $r > 0$. All data are generated on a $256 \times 256$ grid with a pseudospectral method and are downsampled to $32 \times 32$ or $64 \times 64$. The resolution is fixed to $32 \times 32$ for training and $64 \times 64$ for testing.

\subsection{HEAT EQUATION}

The two-dimensional Heat equation for a heated square box on the unit torus takes the form:
$$\partial_{t}u(x,t)=\alpha\Delta u(x,t)+q(x),\quad x\in (0,1)^{2},t\in(0,T],$$
$$u(x,0)=u_{0}(x), \quad x\in (0,1)^{2}.$$

The initial condition $u_{0}(x)$ is generated according to $u_0 \backsim \mu$ where $\mu = \mathcal{N}(0, 7^{3/2}(-\Delta+49I)^{-2.5})$ , with periodic boundary conditions, the heat source $q|_{\partial \Omega} = 0.1$ and the thermal diffusivity is set to $\alpha = 1$e$-4$. Here, Fourier Neural Operators are chosen for solving this equation, learning the operator mapping the vorticity up to time 10 to the solution up to some later time $T > 10$, $G^{\dagger}: C([0,10];H_{per}^{r}((0,1);\mathbb{R})) \to C([10,T];H_{per}^{r}((0,1));\mathbb{R})$ defined by $\omega|_{(0,1)^{2} \times (0,10]} \mapsto \omega|_{(0,1)^{2} \times (10,T]}$ for any $r > 0$. All data are generated on a $256 \times 256$ grid with a pseudospectral method and are downsampled to $32 \times 32$ or $64 \times 64$. The resolution is fixed to be $32 \times 32$ for training and $64 \times 64$ for testing.

\subsection{AIRFOIL}

For this problem, we study the two-dimensional time-dependent compressible flow around the cross-section of airfoils, with different inflow speeds (Mach numbers) and angles of attack, and the NS equation is also used to describe the problem. Here Operators Transformer are chosen for this problem, learning the mapping the velocity up to time $0.576s$ to the solution up to $T = 4.8s$, $G^{\dagger}: \boldsymbol{u}(\cdot,t)|_{t \in [0,0.576]} \mapsto \boldsymbol{u}(\cdot,t)|_{t \in (0.576,4.800]}$. All data on irregular grids are generated by \citep{Mesh-Based}, with the conventional solver SU2 \citep{su2}.

\newpage
\section{Additional Experiment about Hyperparameter Analysis}
\label{Appendix hyperparam}

\begin{figure}[htb]
    \centering
    \includegraphics[width=0.99\linewidth]{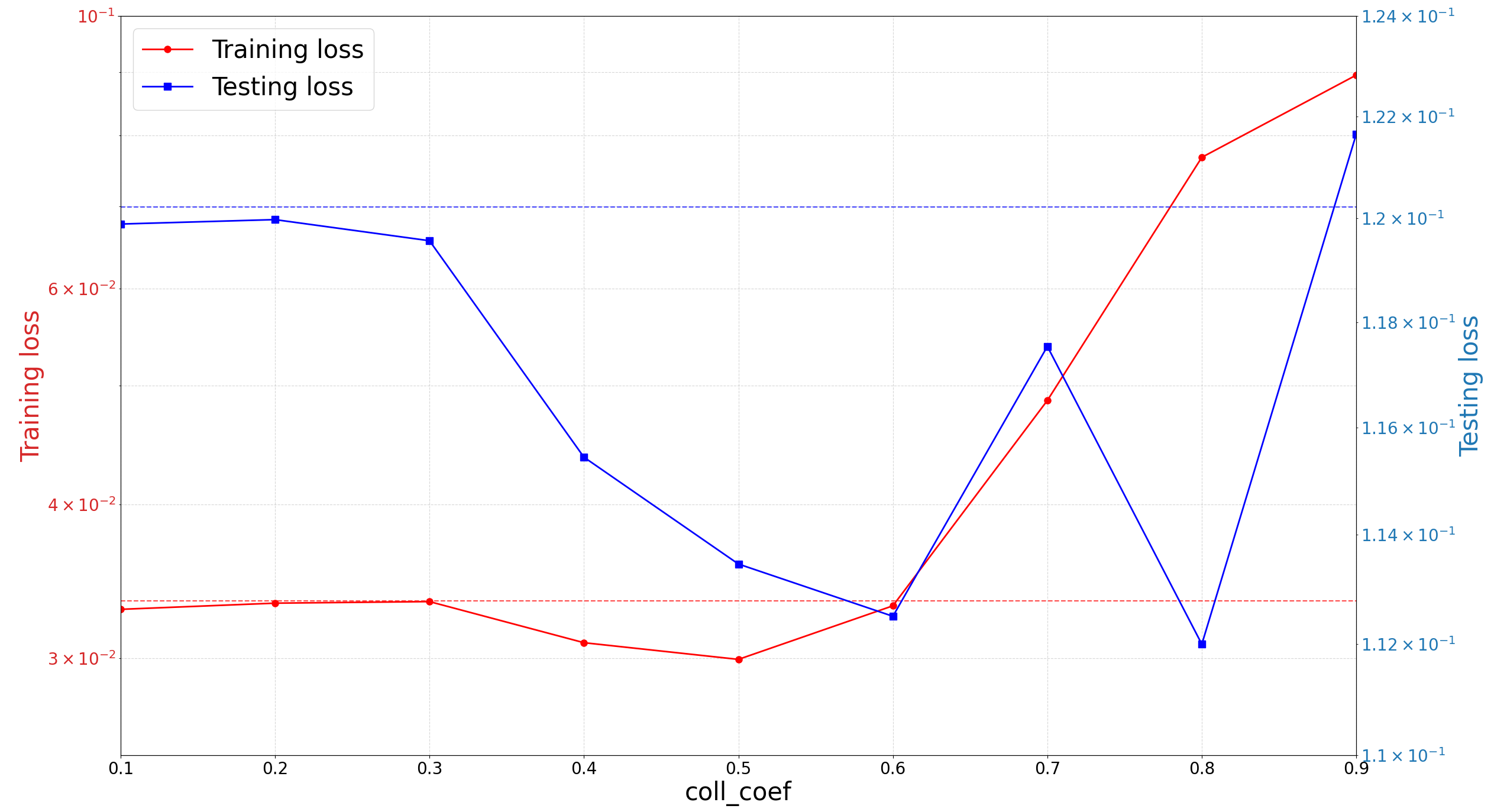}
    \caption{The performance of KITI-FNO with different hyper-parameter  $\text{coll}\_{\textrm{coef}}$ on NS equation. The red and blue dashed lines show the performance of vanilla FNO as baselines.}
    \label{fig:ndivide}
\end{figure}

\section{Additional Experiment about Condensation}
\label{Appendix condensation}

\begin{figure*}[t]
\centering
\begin{subfigure}[t]{0.23\linewidth}
  \centering
  \includegraphics[width=\linewidth]{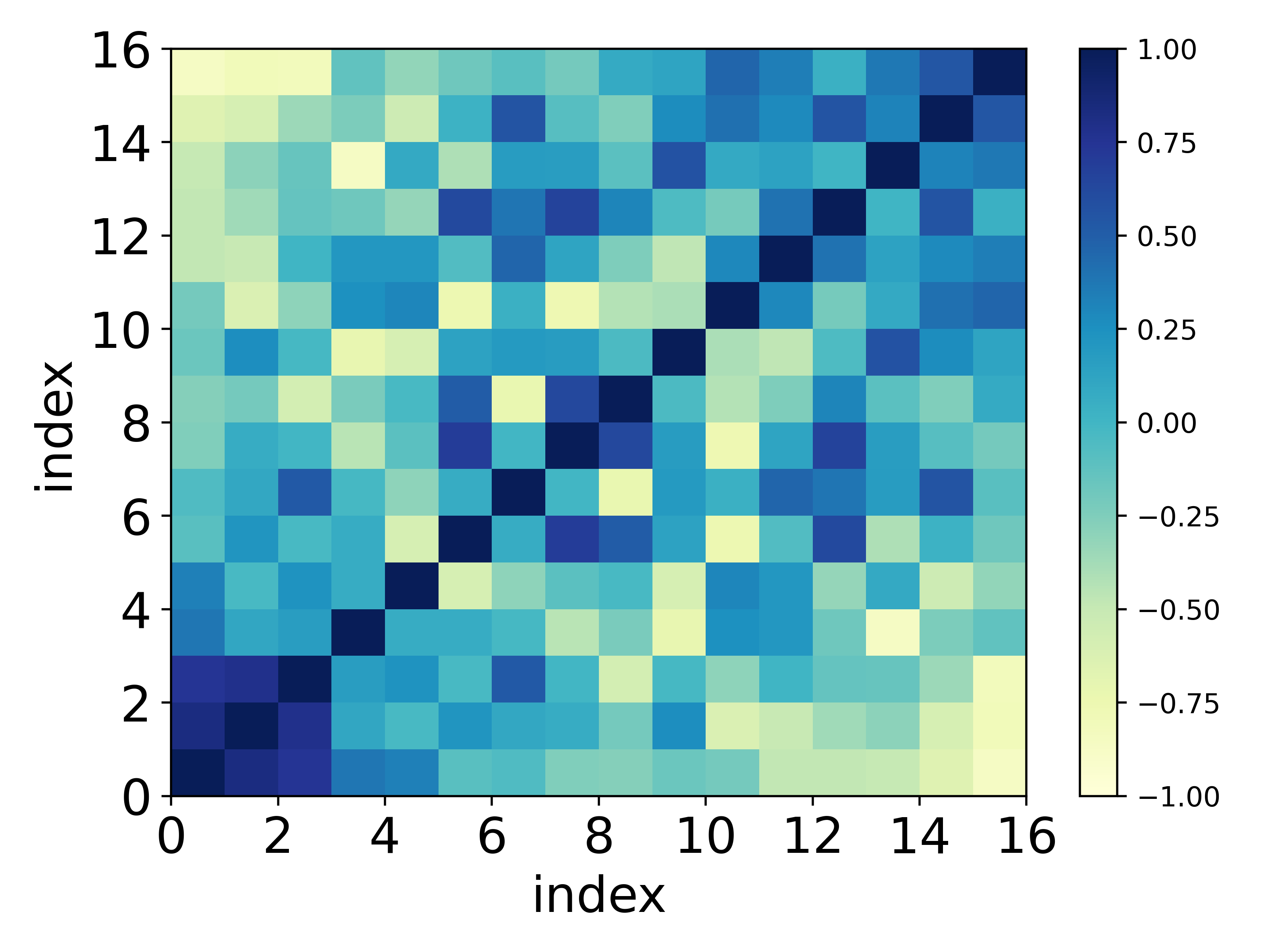}\\
  \includegraphics[width=\linewidth]{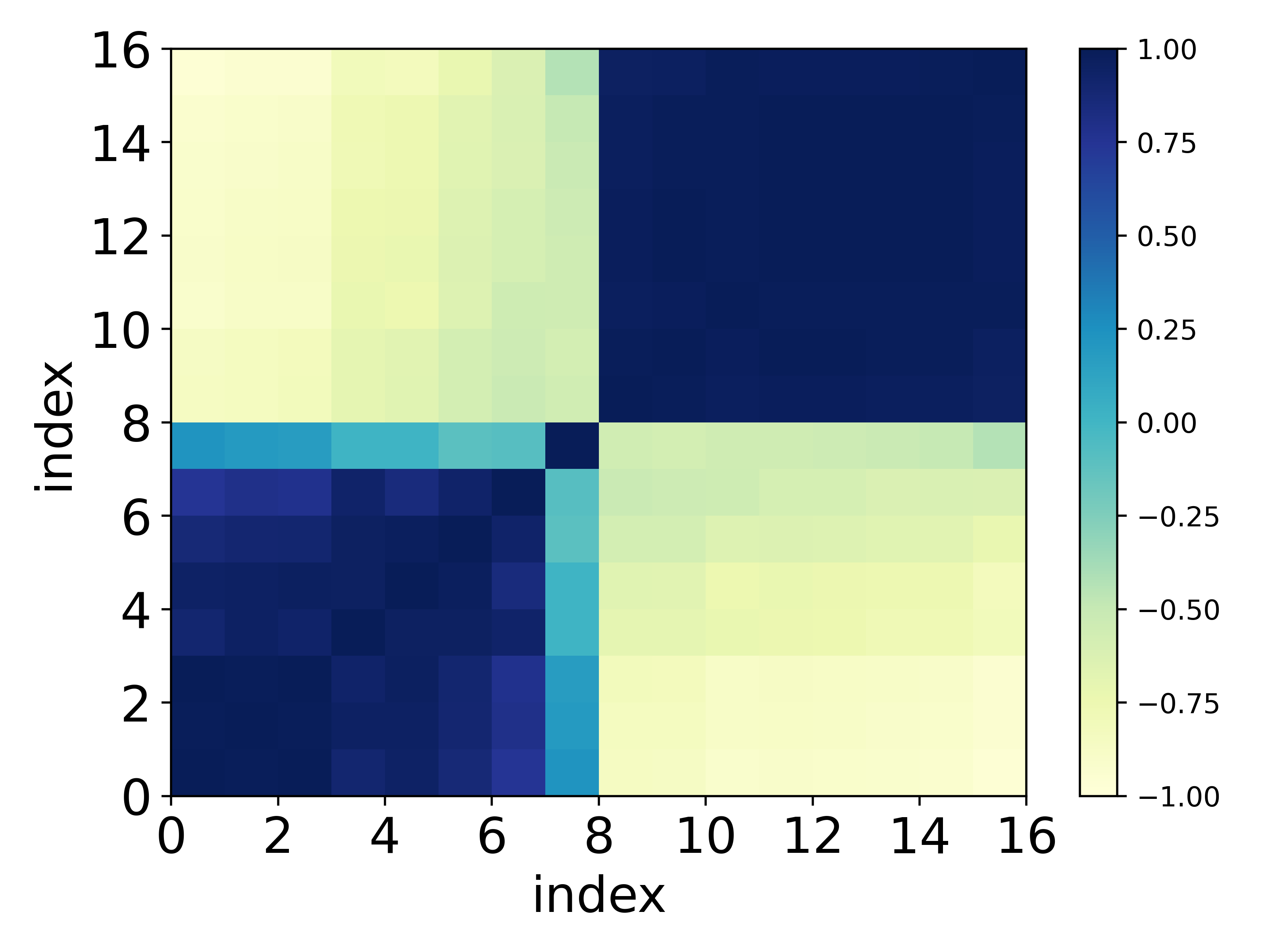}
  \caption{ReLU}
\end{subfigure}
\hfill
\begin{subfigure}[t]{0.23\linewidth}
  \centering
  \includegraphics[width=\linewidth]{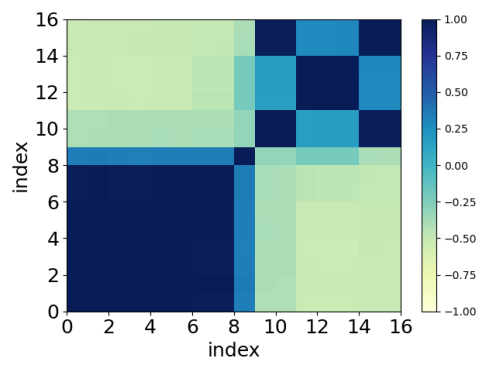}\\
  \includegraphics[width=\linewidth]{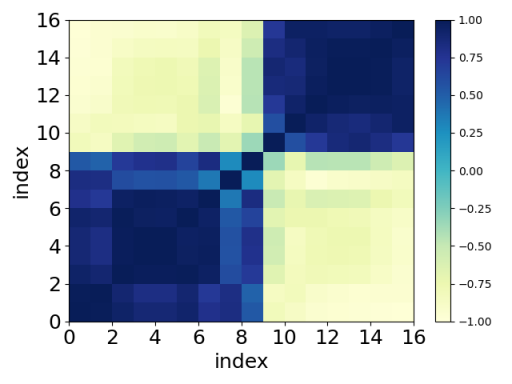}
  \caption{LeakyReLU}
\end{subfigure}
\hfill
\begin{subfigure}[t]{0.23\linewidth}
  \centering
  \includegraphics[width=\linewidth]{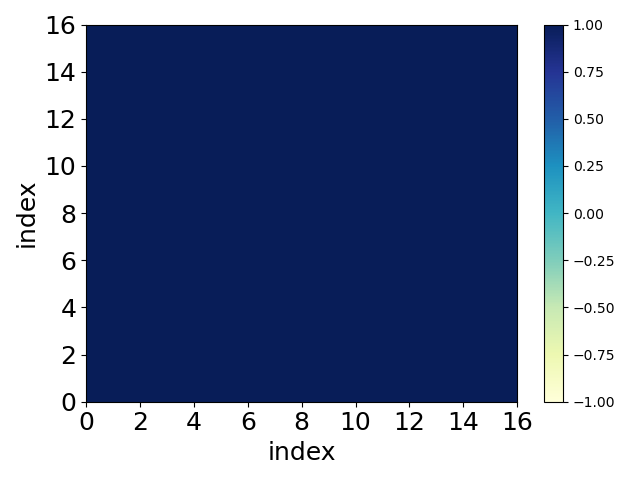}\\
  \includegraphics[width=\linewidth]{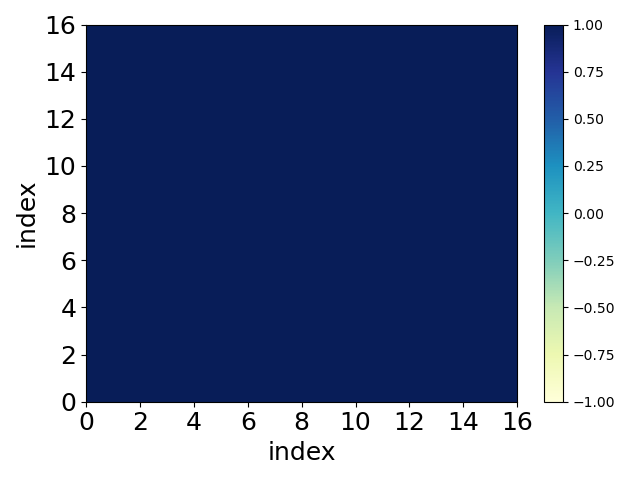}
  \caption{Sigmoid}
\end{subfigure}
\hfill
\begin{subfigure}[t]{0.23\linewidth}
  \centering
  \includegraphics[width=\linewidth]{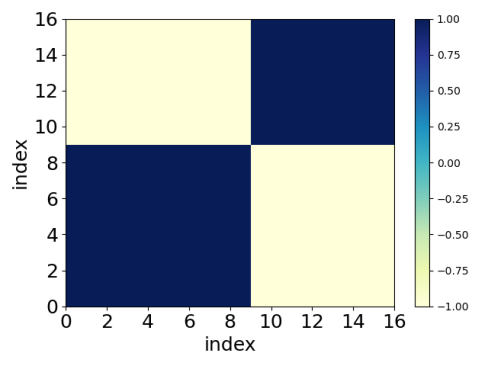}\\
  \includegraphics[width=\linewidth]{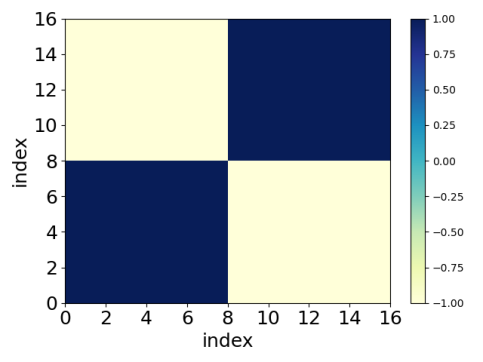}
  \caption{Tanh}
\end{subfigure}
\caption{Results of parameter condensation   on Three-layer Fully-connected Network. (Row 1) linear networks versus (Row 2) KITINet -incorporated networks. Systematic validation is performed across four activation functions: ReLU, LeakyReLU, Sigmoid, and Tanh.}
\label{evolution}
\end{figure*}

\begin{figure*}[t]
\centering
\begin{subfigure}[t]{0.23\linewidth}
  \centering
  \includegraphics[width=\linewidth]{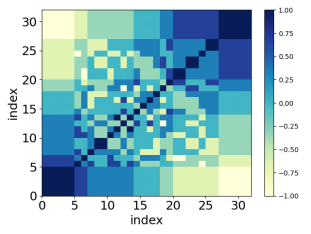}
  \caption{Step 1}
\end{subfigure}
\hfill
\begin{subfigure}[t]{0.23\linewidth}
  \centering
  \includegraphics[width=\linewidth]{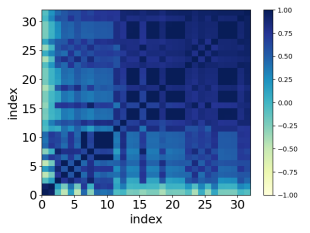}
  \caption{Step 10}
\end{subfigure}
\hfill
\begin{subfigure}[t]{0.23\linewidth}
  \centering
  \includegraphics[width=\linewidth]{condensation_figures/3-layer-sigmoid-pre.png}
  \caption{Step 50}
\end{subfigure}
\hfill
\begin{subfigure}[t]{0.23\linewidth}
  \centering
  \includegraphics[width=\linewidth]{condensation_figures/3-layer-tanh-pre.png}
  \caption{Step 100}
\end{subfigure}
\caption{Evolution of parameter condensation effect on Six-layer ReLU skip-connected network without applying KITINet architecture. The process of parameter condensation is relatively slow.}
\label{fig:evolve}
\end{figure*}

\begin{figure*}[t]
\centering
\begin{subfigure}[t]{0.23\linewidth}
  \centering
  \includegraphics[width=\linewidth]{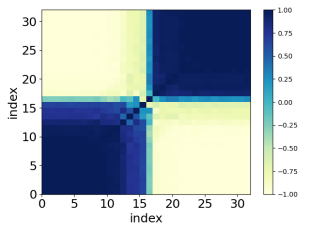}
  \caption{Step 1}
\end{subfigure}
\hfill
\begin{subfigure}[t]{0.23\linewidth}
  \centering
  \includegraphics[width=\linewidth]{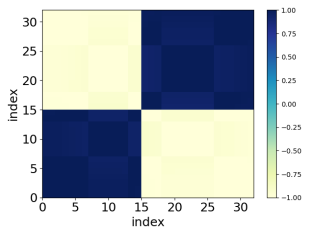}
  \caption{Step 10}
\end{subfigure}
\hfill
\begin{subfigure}[t]{0.23\linewidth}
  \centering
  \includegraphics[width=\linewidth]{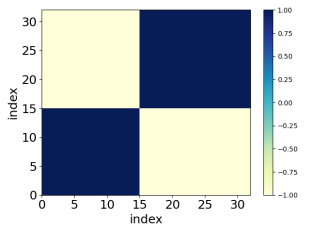}
  \caption{Step 50}
\end{subfigure}
\hfill
\begin{subfigure}[t]{0.23\linewidth}
  \centering
  \includegraphics[width=\linewidth]{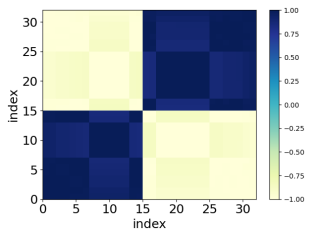}
  \caption{Step 100}
\end{subfigure}
\caption{Evolution of parameter condensation effect on Six-layer ReLU skip-connected network applying KITINet architecture on the last layer. The process of parameter condensation is relatively faster.}
\end{figure*}

\begin{figure*}[t]
\centering
\begin{subfigure}[t]{0.23\linewidth}
  \centering
  \includegraphics[width=\linewidth]{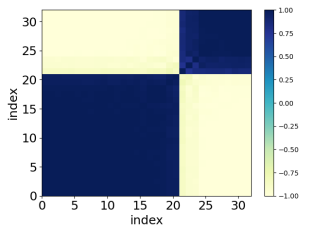}
  \caption{Step 1}
\end{subfigure}
\hfill
\begin{subfigure}[t]{0.23\linewidth}
  \centering
  \includegraphics[width=\linewidth]{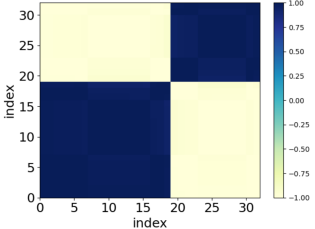}
  \caption{Step 10}
\end{subfigure}
\hfill
\begin{subfigure}[t]{0.23\linewidth}
  \centering
  \includegraphics[width=\linewidth]{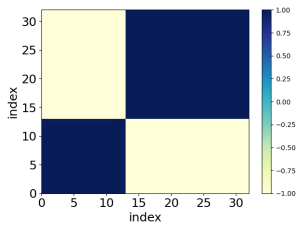}
  \caption{Step 50}
\end{subfigure}
\hfill
\begin{subfigure}[t]{0.23\linewidth}
  \centering
  \includegraphics[width=\linewidth]{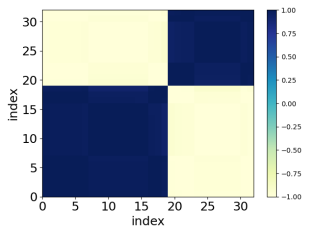}
  \caption{Step 100}
\end{subfigure}
\caption{Evolution of parameter condensation on Six-layer skip-connected Network applying KITINet architecture on the last two layers. The process of parameter condensation is significantly faster and more stable.}
\end{figure*}

\section{The Formal Definition of the Condensation Degree.}
\label{Sec: difinition of degree}
In this section, we provide the formal definitions of condensation for fully connected networks and convolutional networks, which are given in Definitions \ref{def1} and \ref{def2}.

\begin{definition}
    [Weight Correlation in FCN]
Given weight matrix $\nc_\lc \in \mathbb{R}^{N_{\lc-1} \times N_\lc}$ of the $\lc$-th layer, the average weight correlation is defined as
\begin{equation}
    \rho(\nc_\lc) =\frac{1}{N_\lc(N_\lc-1)} \sum_{i,j=1 \atop i \ne j}^{N_\lc} \frac{|\nc_{\lc i}^T \nc_{\lc j}|}{||\nc_{\lc i}||_2 ||\nc_{\lc j}||_2},
\end{equation}
where $\nc_{\lc i}$ and $\nc_{\lc j}$ are $i$-th and $j$-th column of the matrix $\nc_\lc$, corresponding to the $i$-th and $j$-th neuron at $\lc$-th layer, respectively.  Intuitively, $\rho(\nc_\lc)$ is the average cosine similarity between weight vectors of any two neurons at the $\lc$-th layer. 
\label{def1}
\end{definition}
%\lz{I think intuition and example would be good for the definition.
%In Eq(1): $w$ should be used instead of $\wc$? Shall different  letters be chosen for FCN and CNN?} \gaojie{We just want to use $w$ and $\wc$ to differentiate weight matrix in FCN and CNN, does it make sense?}

\begin{definition}
    [Weight Correlation in CNN]
Given the filter tensor $\fc_\lc \in \mathbb{R}^{f \times f \times N_{\lc-1} \times N_\lc}$ of the $\lc$-th layer, where $f \times f$ is the size of the convolution kernel, $\fc_{\lc i} \in \mathbb{R}^{f \times f \times N_{\lc-1}}$ and $\fc_{\lc j} \in \mathbb{R}^{f \times f \times N_{\lc-1}}$ are the $i$-th and $j$-th filter, respectively, of the filter tensor $\fc_\lc$. By reshaping $\fc_{\lc i}$ and $\fc_{\lc j}$ into $\fc'_{\lc i} \in \mathbb{R}^{f^2 \times N_{\lc-1}}$ and $\fc'_{\lc j} \in \mathbb{R}^{f^2 \times N_{\lc-1}}$, respectively, the weight correlation is defined as
\begin{equation}
    \rho(\fc_\lc)=\frac{1}{N_\lc(N_\lc-1)N_{l-1}} \sum_{i,j=1 \atop i \ne j}^{N_\lc}  \sum_{z=1}^{N_{l-1}}\frac{|{\fc'}_{\lc i,z}^T{\fc'}_{\lc j,z}|}{||\fc'_{\lc i,z}||_2 ||\fc'_{\lc j,z}||_2},
\end{equation}
where $\fc'_{\lc i,z}$ and $\fc'_{\lc j,z}$ are the $z$-th column   of $\fc'_{\lc i}$ and $\fc'_{\lc j}$ respectively. Intuitively, $\rho(\fc_\lc)$ is defined as the cosine similarity between filter matrices.
\label{def2}
\end{definition}

\section{\new{A Theory Analysis of Condensation.}}
\label{Appendix: theory}
\subsection{Setting.}
\label{theory setting}
First, we present a formulation of the problem statement and establish the necessary notation that will be used throughout our theoretical analysis. Consider a linear regression problem $y_i = \boldsymbol{k}\boldsymbol{x_i},i=\{1,2,\cdots,n\}$, where the training data $\boldsymbol{x_i}\in \mathbb{R}^d\sim N(0,I)$, $\bold{k}\in\mathbb{R}^{1\times d}$. We use $y=\boldsymbol{a}^\top \boldsymbol{W} \boldsymbol{x}$ to retain the $\boldsymbol{k}$. We set $\boldsymbol{W}$ is a $m\times d$ matrix and $\boldsymbol{a}$ is a $1\times m$ matrix. We use $\boldsymbol{W}(t),\boldsymbol{a}(t)$ to denote the value of $\boldsymbol{W},\boldsymbol{a}$ at step $t$.

Let $\boldsymbol{W}=\begin{bmatrix} \boldsymbol{w}_{1}^\top \\\ \boldsymbol{w}_{2}^\top \\\ \vdots \\\ \boldsymbol{w}_{m}^\top \end{bmatrix}$, $\boldsymbol{w}_{i} \in \mathbb{R}^{d\times 1}$, we assume that  ${w}_{i,j}(t)\sim N(0,\sigma_1(t))$ at step $t$ for every $i\in[m], j\in[d]$. 
Let $\boldsymbol{a} = \begin{bmatrix} a_{1} & a_{2} & \cdots & a_{m} \end{bmatrix}$. we assume that  $a_{i}(t)\sim N(0,\sigma_2(t))$ at step $t$ for every $i\in[m]$.

We consider MSE-loss as follows:

\begin{equation}
    \min_{\boldsymbol{W},\boldsymbol{a}}\ L = \dfrac{1}{2n}\sum_{i=1}^n(\boldsymbol{a}^\top \boldsymbol{W} \boldsymbol{x_i}-y_i)^2
\end{equation}

Let $\boldsymbol{\theta}(t) = \boldsymbol{a}(t)^\top \boldsymbol{W}(t)$ and $\boldsymbol{\theta}^* = \boldsymbol{k}$. We assume that the neural network is over-parametrized, i.e., $m \gg d$. Let $\eta$ denote the learning rate.

\subsection{Case 1: the Circumstance without Collision.}
\begin{theorem}
    \textit{Under the setting in \cref{theory setting}, the convergence rate of the model parameters is exponential.}
\label{thm: case1}
\end{theorem}

\textbf{Proof of \cref{thm: case1}.}

The gradient is
\begin{gather}
    \dfrac{\partial L}{\partial \boldsymbol{w}_{l}^\top} = a_{l}^\top\cdot \dfrac{1}{n}\sum_{i=1}^n(\boldsymbol{a}^\top \boldsymbol{W}\boldsymbol{x}_i - y_i)\cdot \boldsymbol{x}_i^\top \\
     \dfrac{\partial L}{\partial a_{l}^\top} = \boldsymbol{w}_{l}^\top \cdot \dfrac{1}{n}\sum_{i=1}^n(\boldsymbol{a}^\top \boldsymbol{W} \boldsymbol{x}_i - y_i)\cdot \boldsymbol{x}_i^\top 
\end{gather}

By using GD, we have
\begin{gather} 
    \boldsymbol{W}(t+1) = \boldsymbol{W}(t) - \eta\cdot \boldsymbol{a}(t)^\top \cdot \dfrac{1}{n}\sum_{i=1}^n(\boldsymbol{a}^\top(t)\boldsymbol{W}(t)\boldsymbol{x}_i - y_i)\cdot \boldsymbol{x}_i^\top   \\
    \boldsymbol{a}(t+1) = \boldsymbol{a}(t) - \eta\cdot \boldsymbol{W}(t)^\top \cdot \dfrac{1}{n}\sum_{i=1}^n(\boldsymbol{a}(t)^\top \boldsymbol{W}(t)\boldsymbol{x}_i - y_i)\cdot \boldsymbol{x}_i^\top
\end{gather}

By multiplying the above two equations, we get
\begin{align*}
    \boldsymbol{\theta}(t+1)&= \boldsymbol{\theta}(t)-\eta\cdot\boldsymbol{a}(t)\boldsymbol{a}(t)^\top \\&\quad\cdot\dfrac{1}{n}\sum_{i=1}^n(\boldsymbol{a}(t)^\top\boldsymbol{W}(t)\boldsymbol{x}_i-y_i) \\&\qquad\cdot\boldsymbol{x}_i^\top \\
    &\quad-\eta\cdot\dfrac{1}{n}\sum_{i=1}^n(\boldsymbol{a}(t)^\top\boldsymbol{W}(t)\boldsymbol{x}_i-y_i) \\&\qquad\cdot\boldsymbol{x}_i^\top\boldsymbol{W}(t)^\top\boldsymbol{W}(t)+O(\eta^2)
\end{align*}

By the Large Number Law, we have
\begin{gather}
\boldsymbol{a}(t) \boldsymbol{a}(t)^\top= \sum_{i=1}^m a_{i}(t)^2 \approx\sigma_2(t)\cdot m \\ 
\boldsymbol{W}(t)^\top \boldsymbol{W}(t)=\sum_{i=1}^m \boldsymbol{w}_{i}(t)\boldsymbol{w}_{i}(t)^\top \approx\sum_{i=1}^m \sigma_1(t)\cdot I = m\cdot \sigma_1(t)\cdot I
\end{gather}

Since $y_i = \boldsymbol{\theta}^* \cdot\boldsymbol{x}_i$, we can approximately get
\begin{align}
    \boldsymbol{\theta}(t+1)-\boldsymbol{\theta}^*&\approx \boldsymbol{\theta}(t)-\boldsymbol{\theta}^* \notag\\&\quad-\eta\cdot(1+m\cdot(\sigma_1(t)+\sigma_2(t))) \notag\\&\qquad\cdot(\boldsymbol{\theta}(t)-\boldsymbol{\theta}^*)\cdot\dfrac{1}{n}\sum_{i=1}^n \boldsymbol{x}_i\cdot\boldsymbol{x}_i^\top
\end{align}

By the large number theorem, when n is sufficiently large, 
\begin{equation}
    \dfrac{1}{n}\sum_{i=1}^n \boldsymbol{x}_i\cdot \boldsymbol{x}_i^\top \approx E[\boldsymbol{x}_i\cdot \boldsymbol{x}_i^\top] = I
\end{equation}

So finally we retain
\begin{align}
    \boldsymbol{\theta}(t+1)-\boldsymbol{\theta}^*&\approx (1-\eta\cdot(1+m\cdot(\sigma\_1(t)+\sigma\_2(t)))) \notag\\&\qquad\cdot(\boldsymbol{\theta}(t)-\boldsymbol{\theta}^*)\\
    &\approx (\boldsymbol{\theta}(0)-\boldsymbol{\theta}^*) \cdot \prod_{i=0}^t \beta(t) 
\end{align}
where $\beta(t) = 1-\eta\cdot(1+m\cdot(\sigma_1(t)+\sigma_2(t)))$.

Therefore, under collision-free conditions, the model can rapidly descend to convergence in exponential time as long as $\eta$ is small and properly selected.  \quad\quad\(\square\)

\subsection{Case 2: the Circumstance with Collision (KITINet).}

Our key assumption here is that KITINet is in thermal equilibrium. In this idealized state, the temperature is uniform and constant within the system, and all particles have the same velocity distribution and the same probability to collide. Here, we inject uniform constant noise into $a$ as the simulation for thermal equilibrium. While this simplification differs from the real collision dynamics, it can provide valuable insights into how KITINet enhances model robustness.

\begin{theorem}
    \textit{Under the setting in \cref{theory setting} and assuming KITINet is under thermal equilibrium, the introduction of KITINet collisions changes the convergence process to a two-phase process: \\
(1) The norm of the neuron in $W_1$ first decays to a small scale, inducing rapid reorientation in the low-weight regime. \\
(2) The model converges to a sparse solution through a condensation-like dynamics. }
\label{thm: case2}
\end{theorem}

\textbf{Proof of \cref{thm: case2}.} 

The second phase, in which the convergence behavior under small-weight regimes has been extensively studied ~\citep{Maennel2018GradientDQ, Lyu2021GradientDO, zhou2022towards, Min2023EarlyNA, Ma2024ConvergenceOG}, so we only provide the proof in the first phase.

The update rule to simulate the thermal equilibrium KITINet collision is modeled as follow: for every neuron $\boldsymbol{w}_i$ and $a_i$ at step $t$, we have
\begin{gather}
    \label{iter_w1_eta}
    \begin{aligned}\boldsymbol{w}_{i}(t+1)&=\boldsymbol{w}_{i}(t)-\eta\cdot a_i(t)\cdot\dfrac{1}{n} \\&\quad\cdot\sum_{i=1}^n(\boldsymbol{a}(t)^\top\boldsymbol{W}(t)\boldsymbol{x}_i-y_i)\cdot\boldsymbol{x}_i\end{aligned}\\
    \label{iter_w2_eta}
    a_{i}(t+1) = a_{i}(t)  + \delta_i(t)
\end{gather}
where 
\begin{equation}
    \delta_i(t)=\left\{
\begin{aligned}
& -\eta^{0.25} \quad \text{if $a_{i}(t)= \eta^{0.25}$}\\
& \eta^{0.25} \quad\quad \text{if $a_{i}(t)= -\eta^{0.25}$} \\
& \text{$\sim\{-\eta^{0.25}, \eta^{0.25}\}$}
\ \ \ \ \text{if $a_{i}(t)= 0$}
\end{aligned}
\right.
\end{equation}
Besides, we use a new initialization for $\boldsymbol{a}$: for every $i\in [m]$, we have 
\begin{equation}
    a_i(0)=\left\{
\begin{aligned}
& -\eta^{0.25} \quad \text{with probability 1/4}\\
& \eta^{0.25} \ \quad\quad \text{with probability 1/4} \\
& 0\quad\quad\quad\ \ \  \text{with probability 1/2}
\end{aligned}
\right.
\end{equation}
Notice that $a_i(0)$ follows the stationary distribution. Since 
the transition matrix of $a_i$ is \[
Q = \begin{bmatrix}
0 & 1 & 0  \\
\frac{1}{2} & 0 & \frac{1}{2} \\
0 & 1 & 0 \\
\end{bmatrix}
\]
It is easy to verify that $[1/4,1/2,1/4] = [1/4,1/2,1/4]\cdot Q$.

Thus by large number law, we can approximate $\Vert\boldsymbol{a}(t)\Vert^2$ by
\begin{equation}
        \begin{aligned}\Vert\boldsymbol{a}(t)\Vert^2&=\sum_{i=1}^m |a_i(t)|^2 =m\cdot\mathbb{E}[|a_i(t)|^2] \\&=\dfrac{m\cdot\eta^{0.5}}{2}\end{aligned}
\end{equation}
\begin{lemma}
    Using Algorithm 2, we have \begin{equation}
    \mathbb{E}[\boldsymbol{\theta}(t)] =\mathbb{E}[\boldsymbol{\theta}(0)]
    \label{expectation of theta(t)}
\end{equation}
\end{lemma}
\textbf{Proof.} Notice that
\begin{gather} 
    \begin{aligned}\boldsymbol{W}(t+1)&=\boldsymbol{W}(t+1)-\eta\cdot\boldsymbol{a}(t)\cdot\dfrac{1}{n} \\&\quad\cdot\sum_{i=1}^n(\boldsymbol{a}(t)^\top\boldsymbol{W}(t)\bm{x_i}-y_i)\cdot\bm{x_i^\top}\end{aligned}\\
    \boldsymbol{a}(t+1) = \boldsymbol{a}(t) + \bm{\delta}(t)^\top
\end{gather}

Where $\bm{\delta}(t) = [\delta_1(t),\delta_2(t),\cdots,\delta_m(t)]^\top$.

Notice that $\boldsymbol{a}(t+1)^\top\cdot\boldsymbol{a}(t) = 0$, then we have
\begin{align}
    \boldsymbol{\theta}(t+1) &= \boldsymbol{a}(t+1)^\top\cdot \boldsymbol{W}(t+1) \\
    &=(\boldsymbol{a}(t)^\top+\bm{\delta}(t))\cdot\bigl(\boldsymbol{W}(t) \notag\\&\quad-\eta\cdot\boldsymbol{a}(t)\cdot\dfrac{1}{n}\sum_{i=1}^n(\boldsymbol{a}(t)^\top\boldsymbol{W}(t)\boldsymbol{x}_i-y_i) \notag\\&\qquad\cdot\boldsymbol{x}_i^\top\bigr)\\
    &=\boldsymbol{\theta}(t)+\bm{\delta}(t)\cdot\boldsymbol{W}(t) \notag\\&\quad-(\boldsymbol{a}(t+1)^\top\boldsymbol{a}(t))\cdot\eta\cdot\dfrac{1}{n} \notag\\&\qquad\cdot\sum_{i=1}^n(\boldsymbol{a}(t)^\top\boldsymbol{W}(t)\boldsymbol{x}_i-y_i)\cdot\boldsymbol{x}_i^\top\\
    &= \boldsymbol{\theta}(t) + \bm{\delta}(t)\cdot \boldsymbol{W}(t) \label{theta_equation}
\end{align}
Since $\mathbb{E}[\bm{\delta}(t)] = \bm{0}$, we have
\begin{align*}
        \mathbb{E}[\boldsymbol{\theta}(t)] &= \mathbb{E}[\boldsymbol{\theta}(0)] 
\end{align*}
\(\square\)
\subsubsection{Progressive diminishing of norm using Algorithm 2}

Since $\boldsymbol{x}_i\sim N(0,I)$ i.i.d, by law of large numbers, $\frac{1}{n}\sum_{i=1}^n\boldsymbol{x}_i\cdot \boldsymbol{x}_i^\top$ is approximately to $\mathbb{E}[\boldsymbol{x}_i\cdot\boldsymbol{x}_i^\top] = I$. So we have 
\begin{gather}
    \bm{w_{i}}(t+1) = \bm{w_{i}}(t) - \eta\cdot a_{i}(t)\cdot(\boldsymbol{\theta}(t) - \boldsymbol{\theta}^*)^\top \\
    a_{i}(t+1) = a_{i}(t)  + \delta_i(t)
\end{gather}

Square both sides of the equation, and we have
\begin{align}
    \Vert\boldsymbol{w}_{i}(t+1)\Vert^2&=\Vert\boldsymbol{w}_{i}(t)\Vert^2 \notag\\&\quad-2\eta\cdot a_i(t)\cdot(\boldsymbol{\theta}(t)-\boldsymbol{\theta}^*)^\top\boldsymbol{w}_i(t+1) \notag\\&\quad+\eta^2\cdot a_i(t)^2\Vert\boldsymbol{\theta}(t)-\boldsymbol{\theta}^*\Vert^2 
\end{align}

Since $|a_{i}(t)|\leq \eta^{0.25}$, we have 
\begin{align}
    \dfrac{1}{m}\sum_{i=1}^m\Vert\boldsymbol{w}_i(t+1)\Vert^2&=\dfrac{1}{m}\sum_{i=1}^m\Vert\boldsymbol{w}_i(t)\Vert^2 \notag\\&\quad-2\eta\cdot(\boldsymbol{\theta}(t)-\boldsymbol{\theta}^*)^\top\cdot\dfrac{1}{m}\sum_{i=1}^m a_i(t)\boldsymbol{w}_i(t) \notag\\&\quad+\eta^2\cdot\dfrac{1}{m}\sum_{i=1}^m a_i(t)^2\Vert\boldsymbol{\theta}(t)-\boldsymbol{\theta}^*\Vert^2\\
    &=\dfrac{1}{m}\sum_{i=1}^m\Vert\boldsymbol{w}_i(t)\Vert^2-\dfrac{2\eta}{m}(\boldsymbol{\theta}(t)-\boldsymbol{\theta}^*)^\top\boldsymbol{\theta}(t) \notag\\&\quad+\eta^2\cdot\dfrac{1}{m}\sum_{i=1}^m a_i(t)^2\Vert\boldsymbol{\theta}(t)-\boldsymbol{\theta}^*\Vert^2\\
    &\leq\dfrac{1}{m}\sum_{i=1}^m\Vert\boldsymbol{w}_i(t)\Vert^2-\dfrac{2\eta}{m}(\boldsymbol{\theta}(t)-\boldsymbol{\theta}^*)^\top\boldsymbol{\theta}(t) \notag\\&\quad+\eta^2(\eta^{0.25})^2\Vert\boldsymbol{\theta}(t)-\boldsymbol{\theta}^*\Vert^2\\
    &=\dfrac{1}{m}\sum_{i=1}^m\Vert\boldsymbol{w}_i(t)\Vert^2-\left(\dfrac{2\eta}{m}-\eta^{2.5}\right)\Vert\boldsymbol{\theta}(t)-\boldsymbol{\theta}^*\Vert^2 \notag\\&\quad-\dfrac{2\eta}{m}(\boldsymbol{\theta}(t)-\boldsymbol{\theta}^*)^\top\boldsymbol{\theta}^*\label{relation}
\end{align}

\begin{lemma}
    If we set $a_{i}(0)$ to stationary distribution, then for any $t,k\in\mathbf{N}^+$, we have
\begin{equation}
\mathbf{E}[\delta_i(t)\cdot\delta_i(t+k)] =\left\{
\begin{aligned}
& -\dfrac{\eta^{0.5}}{2}\quad \text{if $k=1$}\\
& 0  \quad\quad\quad\quad \text{if $k>1$}
\end{aligned}
\right.
\label{Markov Stationary}
\end{equation}
\end{lemma}

\textbf{Proof.} Without loss of generality, assume we know that $\delta_i(t) = \eta^{0.25}$, then the distribution changes to $(0,\frac{1}{2},\frac{1}{2})$. So we have \[\begin{aligned}\mathbb{E}[\delta_i(t)\delta_i(t+1)]&=\dfrac{1}{2}\eta^{0.25}(-\eta^{0.25}) \\&\quad+\dfrac{1}{2}\cdot0=-\dfrac{\eta^{0.5}}{2}\end{aligned}\]

Then just after that, the distribution changes from $(0,\frac{1}{2},\frac{1}{2})$ to stationary distribution $(\frac{1}{4},\frac{1}{2},\frac{1}{4})$ again. Thus for any $k>1$ we have
$$\mathbb{E}[\delta_i(t)\cdot \delta_i(t+k)] = 0 $$
\(\square\)

The following lemma reveals the process generated in Phase (1), thereby providing a proof for \cref{thm: case2}.

\begin{lemma}[Progressively diminishing under simulation setup]\label{algorithm 2 decay}
     Under the setting in \cref{theory setting} and assuming KITINet is in thermal equilibrium, there exists a step $t_0\leq \frac{1}{\eta^2}$ such that
     \begin{equation}
        \mathbb{E}[\dfrac{1}{m}\sum_{i=1}^m \Vert \boldsymbol{w}_{i}(t_0)\Vert^2] \leq\sqrt{\eta}
    \end{equation}
\end{lemma}

\textbf{Proof. }
 Firstly, we give a lower bound of $\mathbb{E}[\Vert \boldsymbol{\theta}(t)-\boldsymbol{\theta}^*\Vert^2]$, which shows the decrease of each step.
According to \cref{theta_equation}, we retain
\begin{equation}
    \begin{aligned}\boldsymbol{\theta}(t)-\boldsymbol{\theta}^*&=\boldsymbol{\theta}(0)-\boldsymbol{\theta}^* \\&\quad+\sum_{i=0}^{t-1}\bm{\delta}(i)\cdot\boldsymbol{W}(i)\end{aligned} 
\end{equation}

And Since $\mathbb{E}[\bm{\delta}(i)] = 0$, we have
\begin{equation}
    \begin{aligned}&\mathbb{E}\!\left[(\boldsymbol{\theta}(0)-\boldsymbol{\theta}^*)\cdot\sum_{i=0}^{t-1}\bm{\delta}(i)\boldsymbol{W}(i)\right] \\&\quad=\mathbb{E}[\bm{\delta}(i)]\cdot\mathbb{E}\!\left[(\boldsymbol{\theta}(0)-\boldsymbol{\theta}^*)\cdot\sum_{i=0}^{t-1}\boldsymbol{W}(i)\right] \\&\quad=0\end{aligned}
\end{equation} 

So we have 
\begin{equation}    
    \begin{aligned}\mathbb{E}[\Vert\boldsymbol{\theta}(t)-\boldsymbol{\theta}^*\Vert^2]&=
   \mathbb{E}[\Vert\boldsymbol{\theta}(0)-\boldsymbol{\theta}^*\Vert^2] \\&\quad+(\eta^{0.25})^2\sum_{i=0}^{t-1}\mathbb{E}[\Vert\boldsymbol{W}(i)\Vert^2] \\&\quad+2\sum_{i=0}^{t-1}\sum_{j<i}\mathbb{E}\!\left[(\bm{\delta}(i)\boldsymbol{W}(i))(\bm{\delta}(j)\boldsymbol{W}(j))^\top\right]\label{theta^2}\end{aligned}
\end{equation}

Notice that
\begin{align*}
    &\mathbb{E}\!\left[(\bm{\delta}(i)\boldsymbol{W}(i))(\bm{\delta}(i+1)\boldsymbol{W}(i+1))^\top\right] \\&\quad=\mathbb{E}\!\left[\bm{\delta}(i)(\boldsymbol{W}(i)\boldsymbol{W}(i+1)^\top)\bm{\delta}(i+1)^\top\right]\\
    &=\sum_{1\leq l,r\leq m}\mathbb{E}\!\left[\delta_l(i)\delta_r(i+1)(\boldsymbol{W}(i)\boldsymbol{W}(i+1)^\top)[l,r]\right]\\
    &=\sum_{1\leq l,r\leq m}\mathbb{E}\!\left[\delta_l(i)\delta_r(i+1)\boldsymbol{w}_l(i)^\top\boldsymbol{w}_r(i+1)\right]
\end{align*}

Case 1: $r\neq l$. In this case, $\delta_l$ are independent with $\delta_r$. So we have
\begin{align*}
    &\mathbb{E}[\delta_{l}(i)\delta_{r}(i+1)\cdot\boldsymbol{w}_{l}(i)^\top\boldsymbol{w}_{r}(i+1)] \\
    &=\mathbb{E}\!\left[\delta_l(i)\delta_r(i+1)\boldsymbol{w}_l(i)^\top\bigl(\boldsymbol{w}_r(i) \right.\\&\qquad\left.-\eta\cdot a_r(i)(\boldsymbol{\theta}(i)-\boldsymbol{\theta}^*)^\top\bigr)\right]\\
    &=\mathbb{E}\!\left[\delta_l(i)\delta_r(i+1)\boldsymbol{w}_l(i)^\top\boldsymbol{w}_r(i)\right] \\&\quad-\mathbb{E}\!\left[\delta_l(i)\delta_r(i+1)\boldsymbol{w}_l(i)^\top\eta\cdot a_r(i)(\boldsymbol{\theta}(i)-\boldsymbol{\theta}^*)^\top\right]\\
    &=\mathbb{E}[\delta_l(i)]\cdot\mathbb{E}\!\left[\delta_r(i+1)\boldsymbol{w}_l(i)^\top\boldsymbol{w}_r(i)\right] \\&\quad-\mathbb{E}[\delta_l(i)]\cdot\mathbb{E}\!\left[\delta_r(i+1)\boldsymbol{w}_l(i)^\top\eta\cdot a_r(i)(\boldsymbol{\theta}(i)-\boldsymbol{\theta}^*)^\top\right]\\
    &= 0
\end{align*}

Case 2: $r = l$, by Lemma \ref{Markov Stationary}, we have
\begin{align*}
    &\mathbb{E}[\delta_{l}(i)\delta_{l}(i+1)\cdot\boldsymbol{w}_{l}(i)^\top\boldsymbol{w}_{l}(i+1)] \\
    &=\mathbb{E}\!\left[\delta_l(i)\delta_l(i+1)\boldsymbol{w}_l(i)^\top\bigl(\boldsymbol{w}_l(i) \right.\\&\qquad\left.-\eta\cdot a_l(i)(\boldsymbol{\theta}(i)-\boldsymbol{\theta}^*)^\top\bigr)\right]\\
    &=\mathbb{E}\!\left[\delta_l(i)\delta_l(i+1)\boldsymbol{w}_l(i)^\top\boldsymbol{w}_l(i)\right] \\&\quad-\mathbb{E}\!\left[\delta_l(i)\delta_l(i+1)\boldsymbol{w}_l(i)^\top\eta\cdot a_l(i)(\boldsymbol{\theta}(i)-\boldsymbol{\theta}^*)^\top\right]\\
    &=\mathbb{E}[\delta_l(i)\delta_l(i+1)]\cdot\mathbb{E}[\Vert\boldsymbol{w}_l(i)\Vert^2] \\&\quad-\mathbb{E}\!\left[\delta_l(i)\delta_l(i+1)\boldsymbol{w}_l(i)^\top\eta\cdot a_l(i)(\boldsymbol{\theta}(i)-\boldsymbol{\theta}^*)^\top\right]\\
    &=-\dfrac{\eta^{0.5}}{2}\mathbb{E}[\Vert\boldsymbol{w}_l(i)\Vert^2] \\&\quad-\eta\mathbb{E}\!\left[\delta_l(i)\delta_l(i+1)a_l(i)(\boldsymbol{\theta}(i)-\boldsymbol{\theta}^*)^\top\boldsymbol{w}_l(i)\right]\\
    &= -\dfrac{\eta^{0.5}}{2}\mathbb{E}[\Vert\boldsymbol{w}_{l}(i)\Vert^2] - O(\eta^{1.75})
\end{align*}

Thus we have 

\begin{align*}
%&\mathbb{E}\!\left[(\bm{\delta}(i)\boldsymbol{W}(i))\cdot(\bm{\delta}(i+1)\boldsymbol{W}(i+1))^\top\right]\\&\qquad\cdot(\bm{\delta}(i+1)\boldsymbol{W}(i+1))^\top\Bigr] \\&\quad=\sum_{1\leq l,r\leq m}\mathbb{E}\Bigl[\delta_l(i)\delta_r(i+1)\boldsymbol{w}_l(i)^\top \\&\qquad\cdot\boldsymbol{w}_r(i+1)\Bigr]\\
    &\mathbb{E}\!\left[(\bm{\delta}(i)\boldsymbol{W}(i))\cdot(\bm{\delta}(i+1)\boldsymbol{W}(i+1))^\top\right]\\&\quad=\sum_{1\leq l,r\leq m}\mathbb{E}\!\left[\delta_l(i)\delta_r(i+1)\boldsymbol{w}_l(i)^\top\boldsymbol{w}_r(i+1)\right]\\&=\sum_{l=1}^m\mathbb{E}[\delta_l(i)\delta_l(i+1)]\cdot\boldsymbol{w}_l(i)^\top\boldsymbol{w}_l(i+1)\\ 
    &=-\dfrac{\eta^{0.5}}{2}\sum_{l=1}^m\mathbb{E}[\Vert\boldsymbol{w}_l(i)\Vert^2]-O(m\cdot\eta^{1.75})\\
    &=-\dfrac{\eta^{0.5}}{2}\mathbb{E}[\Vert\boldsymbol{W}(i)\Vert^2]-O(m\cdot\eta^{1.75})
\end{align*}

Then we retain

\begin{align} 
    &\sum_{i=0}^{t-1}\sum_{j<i}\mathbb{E}\!\left[(\bm{\delta}(i)\boldsymbol{W}(i))(\bm{\delta}(j)\boldsymbol{W}(j))^\top\right]\notag\\&\quad=\sum_{i=0}^{t-2}\mathbb{E}\!\left[(\bm{\delta}(i)\boldsymbol{W}(i))(\bm{\delta}(i+1)\boldsymbol{W}_{i+1})^\top\right]\\
    &=-\dfrac{\eta^{0.5}}{2}\sum_{i=0}^{t-2}\mathbb{E}[\Vert\boldsymbol{W}(i)\Vert^2]\notag\\&\quad-O(m\cdot\eta^{1.75})\label{deltaW}
\end{align}

Combining \cref{theta^2} and  \cref{deltaW}, we have

\begin{align*}
    \mathbb{E}[\Vert(\boldsymbol{\theta}(t)-\boldsymbol{\theta}^*)\Vert^2] &=\mathbb{E}[\Vert(\boldsymbol{\theta}(0)-\boldsymbol{\theta}^*)\Vert^2] \\
    &\quad+\eta^{0.5}\sum_{i=0}^{t-1}\mathbb{E}[\Vert\boldsymbol{W}(i)\Vert^2] \\&\quad+2\sum_{i=0}^{t-1}\sum_{j<i}\mathbb{E}\!\left[(\bm{\delta}(i)\boldsymbol{W}(i))(\bm{\delta}(j)\boldsymbol{W}(j))^\top\right] \\&\quad+O(m\cdot\eta^{1.75})\\
    &=\mathbb{E}[\Vert\boldsymbol{\theta}(0)-\boldsymbol{\theta}^*\Vert^2] \\&\quad+\eta^{0.5}\sum_{i=0}^{t-1}\mathbb{E}[\Vert\boldsymbol{W}(i)\Vert^2] \\&\quad-\eta^{0.5}\sum_{i=0}^{t-2}\mathbb{E}[\Vert\boldsymbol{W}(i)\Vert^2]+O(m\cdot\eta^{1.75})\\
    &=\mathbb{E}[\Vert\boldsymbol{\theta}(0)-\boldsymbol{\theta}^*\Vert^2] \\&\quad+\eta^{0.5}\mathbb{E}[\Vert\boldsymbol{W}(t-1)\Vert^2]-O(m\cdot\eta^{1.75})\\
    &\geq \mathbb{E}[\|\boldsymbol{\theta}^\star\|^2] + \eta^{0.5}\mathbb{E}[\Vert \boldsymbol{W}(t-1)\Vert^2] 
\end{align*}

% Since $ \max_{0\leq i\leq t-2}\mathbf{E}[\Vert W_1(i)\Vert^2]$ can be easily bounded by $\Vert W_1(0)\Vert^2$. So $O(\eta^2)$ can be neglected.
% Sum up from 0 to T,  we have:
% \begin{align*}
%  \mathbf{E}[\dfrac{1}{m}\sum_{i=1}^m \Vert \bm{w_{1i}(T+1)}\Vert^2] &\leq \mathbf{E}[\dfrac{1}{m}\sum_{i=1}^m \Vert \bm{w_{1i}(0)}\Vert^2] - (\dfrac{4\eta}{m} - 4\eta^{2.5})\cdot(\dfrac{1 - \beta^{2(T+1)}}{1-\beta^2}\mathbf{E}[\Vert(A(0)-A^*)\Vert^2] +\eta^{0.5}\sum_{i=0}^T\mathbf{E}[\Vert W_1(i)\Vert^2]) \\
%  &\ \ \  - \dfrac{4\eta}{m}\cdot\dfrac{1 - \beta^{T+1}}{1-\beta}\mathbf{E}[A(0)-A^*]\cdot A^* \\
%  &\approx \mathbf{E}[\dfrac{1}{m}\sum_{i=1}^m \Vert \bm{w_{1i}(0)}\Vert^2] - (\dfrac{4\eta}{m} - 4\eta^{2.5})\eta^{0.5}\cdot\sum_{i=0}^T\mathbf{E}[\Vert W_1(i)\Vert^2])\\
%  &\ \ \ - (\dfrac{4\eta}{m} - 4\eta^{2.5})\cdot(\dfrac{\eta^{0.5}}{2}+\Vert A^*\Vert^2)\cdot\dfrac{1}{1-\beta^2} + \dfrac{4\eta}{m}\cdot\dfrac{1}{1-\beta}\cdot\Vert A^*\Vert^2 \quad \\
%  &= \mathbf{E}[\dfrac{1}{m}\sum_{i=1}^m \Vert \bm{w_{1i}(0)}\Vert^2] - (\dfrac{4\eta}{m} - 4\eta^{2.5})\eta^{0.5}\cdot\sum_{i=0}^T\mathbf{E}[\Vert W_1(i)\Vert^2]) + O(\eta/m) \\
%  &\approx \mathbf{E}[\dfrac{1}{m}\sum_{i=1}^m \Vert \bm{w_{1i}(0)}\Vert^2] - \dfrac{4\eta^{1.5}}{m}\cdot\sum_{i=0}^T\mathbf{E}[\Vert W_1(i)\Vert^2]) +  \dfrac{4\eta}{m}\cdot\Vert A^*\Vert \\
% \end{align*}

Then we give a proof of this theorem by contradiction. Assume for every step $t\leq \frac{1}{\eta^2}$, we have 
\begin{equation}
\mathbb{E}[\dfrac{1}{m}\sum_{i=1}^m \Vert \bm{w_{i}}(T)\Vert^2]\geq \sqrt{\eta}
\end{equation}
i.e. 
\begin{equation}
\mathbb{E}[\|\boldsymbol{W}(t)\|^2]\geq m\cdot\sqrt{\eta}
\end{equation}

When $T = \dfrac{1}{\eta^{2}}$, according to \cref{relation}, sum up from $0$ to $T-1$ and we have 
\begin{align*}
 \mathbb{E}\!\left[\dfrac{1}{m}\sum_{i=1}^m\Vert\bm{w}_i(T)\Vert^2\right]&=\mathbb{E}\!\left[\dfrac{1}{m}\sum_{i=1}^m\Vert\boldsymbol{w}_i(0)\Vert^2\right] \\&\quad-\left(\dfrac{2\eta}{m}-\eta^{2.5}\right)\sum_{t=0}^{T-1}\mathbb{E}[\Vert\boldsymbol{\theta}(t)-\boldsymbol{\theta}^*\Vert^2] \\&\quad-\dfrac{2\eta}{m}\sum_{t=0}^{T-1}\mathbb{E}\!\left[(\boldsymbol{\theta}(t)-\boldsymbol{\theta}^*)^\top\boldsymbol{\theta}^*\right]\\ 
 &\leq\mathbb{E}\!\left[\dfrac{1}{m}\sum_{i=1}^m\Vert\boldsymbol{w}_i(0)\Vert^2\right] \\&\quad-\left(\dfrac{2\eta}{m}-\eta^{2.5}\right)\left(T\mathbb{E}[\Vert\boldsymbol{\theta}^*\Vert^2]\right.\\&\qquad+\eta^{0.5}\sum_{t=0}^{T-1}\mathbb{E}[\Vert\boldsymbol{W}_t\Vert^2]\left.\right) \\&\quad+\dfrac{2T\eta}{m}\mathbb{E}[\Vert\boldsymbol{\theta}^*\Vert^2]\\
 &\leq\mathbb{E}\!\left[\dfrac{1}{m}\sum_{i=1}^m\Vert\boldsymbol{w}_i(0)\Vert^2\right] \\&\quad-2\eta^{1.5}T\sqrt{\eta}+O(T\cdot\eta^{2.5})\\
 &=1-2\eta^{1.5}\dfrac{1}{\eta^2}\sqrt{\eta}+O(T\cdot\eta^{2.5}) \\&=-1+O(T\cdot\eta^{2.5})<0
\end{align*}
Which is absolutely a contradiction! Therefore, we complete the proof. \quad\(\square\)

\end{document}